\documentclass[10pt]{IEEEtran}
\usepackage{amsmath,epsfig,amsfonts,bm,algorithm,algorithmicx,algpseudocode,color}

\title{Nonparametric Basis Pursuit\\ via Sparse Kernel-based Learning$^\dag$\footnote{$^\dag$Work in this paper was supported by NSF-EARS grant no. AST-1247885; NIH  grant no. 1R01GM104975-01; and the  AFOSR MURI grant no. FA9550-10-1-0567.}}
\author{Juan Andr\'es Bazerque and Georgios B. Giannakis\\Dept. of ECE and Digital Technology Center\\ Univ. of  Minnesota, Minneapolis, MN 55455, USA}
\def\bci{{{\mathbf c_i}}}
\def\bei{{{\mathbf e_i}}}
\def\bbi{{\mathbf b_i}}

\def\bbCi{{\ensuremath{\mathbf{\bar C}_i}}}
\def\vv{\textrm{vec}}
\def\dd{\textrm{Diag}}
\def\Dw{\mathbf D_{{W}}}

\newcommand{\bx}{\mathbf x}

\newcommand{\Bpi}{\mathcal B_{\pi}}
\newcommand{\sumn}{\sum_{n=1}^N}
\newcommand{\sumnz}{\sum_{n\in\mathbb Z}}
\newcommand{\sump}{\sum_{i=1}^P}

\newcommand{\calX}{\mathcal X}
\newcommand{\calY}{\mathcal Y}
\newcommand{\bK}{\mathbf K}
\newcommand{\Kx}{\mathbf K_{\mathcal X}}
\newcommand{\Ky}{\mathbf K_{\mathcal Y}}

\newcommand{\ky}{k_{\mathcal Y}}
\newcommand{\calHx}{\mathcal H_{\mathcal X}}
\newcommand{\calHi}{\mathcal H_{i}}
\newcommand{\calHy}{\mathcal H_{\mathcal Y}}
\newcommand{\gammai}{{\bm \gamma}_{i}}
\newcommand{\sinc}{\textrm{sinc}}
\newcommand{\sto}{\textrm{s. to }}
\newcommand{\bB}{\mathbf B}
\newcommand{\bZ}{\mathbf Z}
\newcommand{\bz}{\mathbf z}
\newcommand{\bW}{\mathbf W}
\newcommand{\bC}{\mathbf C}
\newcommand{\tbB}{\mathbf{\tilde B}}
\newcommand{\tbC}{\mathbf{\tilde C}}

\newcommand{\tr}{\textrm{trace}}
\newcommand{\bmp}{\bm\phi}
\newtheorem{lemma}{Lemma}

\begin{document}
\markboth{IEEE SIGNAL PROCESSING MAGAZINE,  2013 (TO APPEAR)}
\maketitle\maketitle

\begin{abstract}
Signal processing tasks as fundamental as sampling, reconstruction,
minimum mean-square error interpolation and prediction can be
viewed under the prism of reproducing kernel Hilbert spaces.
Endowing this vantage point with contemporary advances in
sparsity-aware modeling and processing, promotes the nonparametric
basis pursuit advocated in this paper as the overarching framework
for the confluence of kernel-based learning (KBL) approaches
leveraging sparse linear regression, nuclear-norm regularization,
and dictionary learning. The novel sparse KBL toolbox goes beyond
translating sparse parametric approaches to their nonparametric
counterparts, to incorporate new possibilities such as multi-kernel
selection and matrix smoothing. The impact of sparse KBL to signal
processing applications is illustrated through test cases from
cognitive radio sensing, microarray data imputation, and network
traffic prediction.
\end{abstract}
\section{Introduction}

Reproducing kernel Hilbert spaces (RKHSs) provide an orderly analytical
framework for nonparametric regression, with the optimal
kernel-based function estimate emerging as the solution of a
regularized variational problem \cite{W90}. The pivotal role of RKHS
is further appreciated through its connections to ``workhorse''
signal processing tasks, such as the Nyquist-Shannon sampling and reconstruction result that involves  sinc kernels \cite{NS12}. Alternatively, spline kernels  replace  sinc kernels, when smoothness rather than bandlimitedness is to be present in the underlying function space~\cite{U90}.

Kernel-based function estimation can be also seen from a Bayesian viewpoint. RKHS
and linear minimum mean-square error (LMMSE) function estimators coincide
when the pertinent covariance matrix equals the kernel Gram matrix.
This equivalence has been leveraged in the context of field
estimation, where spatial LMMSE estimation referred to as Kriging,
is tantamount to two-dimensional RKHS interpolation \cite{C91}.
Finally, RKHS based function estimators can linked with Gaussian processes (GPs) obtained upon defining their
covariances via kernels \cite{RW06}.

Yet another seemingly unrelated, but increasingly popular theme in
contemporary statistical learning and signal processing, is that of
matrix completion~\cite{F02}, where data organized in a matrix
can have missing entries due to e.g., limitations in the acquisition
process. This article builds on the assertion
that imputing missing entries amounts to interpolation, as in
classical sampling theory, but with the low-rank constraint  replacing that of
bandlimitedness. From this  point of view, RKHS interpolation emerges
as the prudent framework for matrix completion that allows effective
incorporation of a priori information via kernels \cite{ABEV09}, including
sparsity attributes.

Recent advances in sparse signal recovery and regression motivate
a sparse kernel-based learning (KBL) redux, which is the purpose and  core of the present paper.
Building blocks of sparse signal processing include the (group)
least-absolute shrinkage and selection operator (Lasso) and its weighted versions \cite{HTF09}, compressive
sampling \cite{CT05}, and nuclear norm regularization \cite{F02}.
The common denominator behind these operators is the sparsity on a
signal's support that the $\ell_1$-norm regularizer induces. Exploiting sparsity for  KBL leads to several innovations regarding the selection of multiple kernels \cite{MP05,KY10}, additive modeling
\cite{RLLW09,LZ06}, collaborative filtering \cite{ABEV09}, matrix
and tensor completion via dictionary learning \cite{BMG12}, as well
as nonparametric basis selection \cite{BMG11}.  In this context, the main contribution of this paper is a \emph{nonparametric} basis pursuit  (NBP) tool,  unifying and advancing a number of  \emph{sparse} KBL approaches.


Constrained by space limitations, a sample of applications stemming from such an encompassing
analytical tool will be also delineated. Sparse KBL and its
various forms contribute to computer vision \cite{SNPC12,VB02}, cognitive
radio sensing \cite{BMG11}, management of user preferences \cite{ABEV09},
bioinformatics \cite{SL11}, econometrics \cite{LZ06,RLLW09}, and forecasting of electric prices, load, and renewables (e.g., wind speed) \cite{kvlg2013isgt}, to
name a few.

The remainder of the paper is organized as follows. Section II reviews the theory of RKHS in connection with GPs, describing the Representer Theorem and the kernel trick, and presenting the Nyquist-Shannon Theorem (NST) as an example of KBL.  Section III deals with  sparse KBL including sparse additive models (SpAMs) and multiple kernel learning (MKL) as examples of  additive nonparametric models.   NBP is introduced in Section IV, with a basis expansion model capturing the general framework for sparse KBL. Blind versions of NBP  for matrix completion and dictionary learning are developed in Sections V and VI. Finally, Section
VII presents numerical tests  using  real and simulated data, including RF spectrum measurements,  expression levels in yeast,  and network traffic loads. Conclusions are drawn in Section
VIII, while most technical details are deferred to the Appendix.
\section{KBL Preliminaries}\label{KBL-prelims}
In this section, basic tools and approaches are reviewed to place known schemes for nonparametric (function) estimation under a common denominator.
\subsection{RKHS and the Representer Theorem}
In the context of  reproducing kernel Hilbert spaces (RKHS)
\cite{W90}, nonparametric estimation of a function
$f:\calX\rightarrow \mathbb R$ defined over a measurable space
$\calX$ is performed via interpolation of $N$ training points
$\{(x_1,z_1),\ldots,(x_N,z_N)\},$ where $x_n\in \calX,$ and $z_n=f(x_n)+e_n\in \mathbb
R$. For this purpose, a kernel function
$k:\calX\times\calX\rightarrow\mathbb R$  selected to be
\emph{ symmetric}  and  \emph{positive definite,}  specifies a linear space of interpolating functions $f(x)$ given by
\begin{equation}\calHx:\hspace{-0.02cm}=\hspace{-0.02cm}\left\{\hspace{-0.02cm}f(x)\hspace{-0.02cm}=\hspace{-0.08cm}\sum_{n=1}^\infty \hspace{-0.02cm}\alpha_nk(x_n,x)\hspace{-0.02cm}:\alpha_n\hspace{-0.02cm}\in\hspace{-0.02cm}\mathbb R, x_n\hspace{-0.02cm}\in\hspace{-0.02cm}\hspace{-0.02cm}\calX, n\hspace{-0.02cm}\in\hspace{-0.02cm}\mathbb N\hspace{-0.02cm}\right\}\hspace{-0.02cm}. \nonumber
\end{equation}
For many choices of $k(\cdot,\cdot)$, $\calHx$ is exhaustive with respect to (w.r.t) families of  functions obeying certain regularity conditions. The spline kernel for example,  generates the Sobolev space of  all low-curvature functions \cite{D77}. Likewise, the sinc kernel gives rise to the space of bandlimited functions.  Space
$\calHx$ becomes a Hilbert space when equipped with the inner product
$<f,f'>_{\calHx}:=\sum_{n,n'=1}^\infty \alpha_n\alpha'_{n'}k(x_n,x'_{n'})$,
and the associated norm is $\|f\|_{\calHx}:=\sqrt{<f,f>_{\calHx}}$. A key
result in this context  is the so-termed Representer Theorem \cite{W90}, which asserts that
based on $\{(x_n,z_n)\}_{n=1}^N$, the optimal interpolator in $\calHx$, in the sense of
\begin{equation}
\vspace*{-0.2cm}
\hat f=\arg\min_{f\in \calHx} \sumn(z_{n}-f(x_n))^2+\mu
\|f\|^2_{\calHx} \label{representer}
\end{equation}
admits the finite-dimensional representation
\begin{equation}
\hat
f(x)=\sum_{n=1}^N\alpha_nk(x_n,x).
\label{kernel_expansion}
\end{equation}

This result is nice in its simplicity, since functions in space
$\calHx$ are compound by a numerable but arbitrarily large number
of kernels, while $\hat f$ is a combination of just a \emph{finite} number
of kernels around the training points. In addition, the regularizing
term $\mu \|f\|^2_{\calHx}$ controls smoothness, and thus reduces
overfitting.   After substituting \eqref{kernel_expansion} into \eqref{representer}, the coefficients $\bm\alpha^T:=[\alpha_1,\ldots,\alpha_N]$  minimizing the regularized least-squares (LS) cost in  \eqref{representer}  are given by $\bm\alpha=(\mathbf K+\mu\mathbf I)^{-1}\bz$, upon recognizing  that    $\|f\|_{\calHx}^2:=\bm\alpha^T\bK\bm\alpha$, and defining  $\bz^T:=[z_1,\ldots,z_N]$ as well as the kernel dependent Gram matrix $\bK\in\mathbb R^{N\times N}$ with entries $\mathbf K_{n,n'}:=k(x_n,x_{n'})$  ($\cdot^T$ stands for transposition).

\noindent\textbf{Remark 1.} The finite-dimensional expansion \eqref{kernel_expansion} solves \eqref{representer} for more general fitting costs and regularizing terms. In its general form,  the Representer Theorem asserts that \eqref{kernel_expansion} is the solution %
\begin{equation}
\vspace*{-0.2cm}
\hat f=\arg\min_{f\in \calHx} \sumn \ell(z_{n},f(x_n))+\mu \Omega(\|f\|_{\calHx}) \label{general_representer}
\end{equation}
where the loss  function $\ell(z_n, f(x_n))$ replacing the LS cost in \eqref{representer} can be selected to serve either robustness   (e.g., using the absolute-value instead of the square error); or, application dependent objectives (e.g., the Hinge loss  to serve classification applications); or, for  accommodating non-Gaussian noise models when viewing \eqref{general_representer} from a Bayesian angle. On the other hand, the regularization term can be chosen as  any increasing function $\Omega$ of the  norm  $\|f\|_{\calHx},$ which will turn out to be crucial for introducing the notion of sparsity, as described in the ensuing sections.

\subsection{LMMSE, Kriging,  and GPs}
Instead of the deterministic treatment of the previous subsection,  the unknown $f(x)$ can be considered as a random process. The KBL estimate \eqref{kernel_expansion} offered by the Representer Theorem has been linked with the LMMSE-based estimator of random fields $f(x)$, under the term Kriging \cite{C91}.
 To predict the value  $\zeta=f(x)$ at an exploration point $x$ via Kriging, the predictor $\hat f(x)$ is modeled as a linear combination of noisy samples  $z_n:=f(x_n)+\eta(x_n)$ at  measurement points $\{x_n\}_{n=1}^N$; that is,
\begin{equation}\label{lmmse}
  \hat f(x)=\sumn \hat\beta_n z_n = \mathbf z^T\bm{\hat \beta}
\end{equation}
where $\bm{\hat \beta}^T:=[\hat\beta_1,\ldots,\hat\beta_N]$ are the expansion coefficients,  and $\mathbf z^T:=[z_1,\ldots,z_N]$ collects the data. The MSE criterion is adopted to find the optimal $\bm{\hat \beta}:=\arg\min_{\bm\beta} E[f(x)-\mathbf z^T \bm \beta]^2$. Solving the latter yields  $\bm{\hat\beta}=\mathbf R_{\bz\bz}^{-1}\mathbf r_{\bz \zeta}$, where $\mathbf R_{\bz\bz}:=E[\bz\bz^T]$ and $\mathbf r_{\bz \zeta}:=E[\mathbf z f(x)]$. If $\eta(x)$ is zero-mean white noise with power $\sigma_\eta^2$, then $\mathbf R_{\bz\bz}$ and $\mathbf r_{\mathbf z\zeta}$ can be expressed in terms of the unobserved $\bm\zeta^T:=[f(x_1),\ldots,f(x_N)]$ as $\mathbf R_{\bz\bz}=\mathbf R_{\bm \zeta\bm\zeta}+\sigma_{\eta}^2\mathbf I$, where $\mathbf R_{\bm\zeta\bm\zeta}:=E[\bm\zeta\bm\zeta^T]$,  and $\mathbf r_{\mathbf z\zeta}=\mathbf r_{\bm\zeta\zeta}$, with $\mathbf r_{\bm\zeta\zeta}:=E[\bm\zeta f(x)]$. Hence, the LMMSE estimate  in \eqref{lmmse} takes the form
\begin{equation}\label{kriging}
\hat f(x)=\mathbf z^T(\mathbf R_{\bm\zeta\bm\zeta}+\sigma_{\eta}^2\mathbf I)^{-1}\mathbf r_{\bm\zeta\zeta}=\sumn \alpha_n r(x,x_n)
\end{equation}
where   $\bm\alpha^T:=\bz^T(\mathbf R_{\bm\zeta\bm\zeta}+\sigma_{\eta}^2\mathbf I)^{-1},$ and the $n$-th entry of $\mathbf r_{\bm\zeta \zeta}$, denoted by $r(x_n, x):= E[f(x)f(x_n)]$, is indeed a function of  the
exploration point $x$, and the measurement point $x_n$.

With the Kriging estimate given by \eqref{kriging}, the RKHS and LMMSE estimates  coincide when the kernel in \eqref{kernel_expansion} is
chosen equal to the covariance function $r(x,x')$ in \eqref{kriging}.

The linearity assumption in \eqref{lmmse} is unnecessary when  $f(x)$ and $e(x)$ are modeled as  zero-mean GPs \cite{RW06}.  GPs are those in which instances of the field at arbitrary points are jointly Gaussian. Zero-mean GPs are specified by  $\textrm{cov}(x,x'):=E[f(x)f(x')]$, which determines the covariance matrix of any vector comprising  instances of the field, and thus its specific zero-mean Gaussian distribution. In particular, the vector $\bm{\bar \zeta}^T:=[f(x), f(x_1), \ldots,f(x_N)]$ collecting the field at the exploration and measurement points is Gaussian, and so  is the vector  $\mathbf{\bar z}^T:=[f(x), f(x_1)+\eta(x_1), \ldots,f(x_N)+\eta(x_N)]=[\zeta,\bz^T].$ Hence, the MMSE estimator,  given by the expectation of $f(x)$ conditioned on $\bz$,  reduces to \cite{K01}
\begin{equation}\label{conditional_mean}
\hat f(x)=E(f(x)|\mathbf z)=\bz^T \mathbf  R_{\bz\bz}^{-1} \mathbf r^T_{\bz \zeta} =\sumn \alpha_n \textrm{cov}(x_n,x).
\end{equation}

By comparing   \eqref{conditional_mean} with \eqref{kriging}, one deduces that the MMSE estimator of a GP coincides with the LMMSE estimator, hence with the RKHS estimator, when $\textrm{cov}(x,x')=k(x,x')$.

\subsection{The kernel trick}
Analogous to the spectral decomposition of matrices,  Mercer's Theorem establishes that if the  symmetric positive definite kernel is square-integrable, it admits a possibly infinite eigenfunction decomposition $k(x,x')=\sum_{i=1}^{\infty} \lambda_i e_i(x)e_i(x')$  \cite{W90}, with  $<e_i(x),e_{i'}(x)>_{\calHx}=\delta_{i-i'}$ where $\delta_{i}$ stands for  Kronecker's delta. Using the weighted eigenfunctions $\phi_i(x):=\sqrt{\lambda_i}e_i(x),\ i\in \mathbb N,$ a point $x\in \calX$ can be mapped to a vector (sequence) $\bm\phi\in\mathbb R^{\infty}$ such that $\phi_i=\phi_i(x),\ i \in \mathbb N.$ This mapping interprets  a kernel as an inner product in $\mathbb R^\infty$, since for two points $x,x'\in\calX,$ $k(x,x')=\sum_{i=1}^{\infty}\phi_i(x)\phi_i(x'):=\bm\phi^T(x)\bm\phi(x')$. Such an inner product interpretation forms  the basis for the \emph{``kernel trick.''}

The kernel trick allows for approaches that depend on inner products of functions (given by infinite kernel expansions) to be recast and implemented using finite dimensional covariance (kernel) matrices. A simple demonstration of this valuable property can be provided through kernel-based ridge  regression. Starting from the standard ridge estimator $\bm{\hat\beta}:=\arg\min_{\bm\beta\in \mathbb R^D} \sumn (z_n-\bmp_n^T\bm  \beta)^2+\mu\|\bm\beta\|^2$ for $\bmp_n \in \mathbb R^D$, and $\bm\Phi:=[\bm\phi_1,\ldots,\bm\phi_N]$, it is possible to rewrite and solve  $\bm{\hat\beta}= \arg\min_{\bm\beta\in \mathbb R^D}\|\bz -\bm\Phi^T\bm\beta\|^2+\mu\|\bm\beta\|^2=(\bm\Phi\bm\Phi^T+\mu\mathbf I)^{-1}\bm\Phi\bz$.  After $\bm{\hat\beta}$ is obtained in the training phase, it can be used for prediction of an ensuing $\hat z_{N+1}=\bm\phi_{N+1}^T\bm{\hat\beta}$ given $\bm\phi_{N+1}$. By using the matrix inversion lemma, $\hat z_{N+1}$ can be written as $\hat z_{N+1}=(1/\mu) \bmp_{N+1}^T\bm\Phi\bz-(1/\mu)\bmp_{N+1}^T\bm\Phi(\mu\bm I+\bm\Phi^T\bm\Phi)^{-1}\bm\Phi^T\bm\Phi\bz$.

Now, if $\bm\phi_n=\bm\phi(x_n)$ with $D=\infty$ is constructed from  $x_n\in \calX$ using eigenfunctions $\{\phi_i(x_n)\}_{i=1}^{\infty}$,  then  $\bmp^T_{N+1}\bm\Phi =\mathbf k^T(x_{N+1}):=[k(x_{N+1},x_1),\ldots,k(x_{N+1},x_N) ],$ and $\bm\Phi^T\bm\Phi=\bK$, which yields
\begin{align}
\nonumber \hat z_{N+1}&=(1/\mu) \mathbf k^T(x_{N+1})[\mathbf I- (\mu\bm I+\bK)^{-1}\bK]\bz\\
&=\mathbf k^T(x_{N+1})(\mu\bm I+\bK)^{-1}\bz\label{kernel_trick}
\end{align}
coinciding with \eqref{conditional_mean}, \eqref{kriging}, and with the solution of \eqref{representer}.

Expressing  a linear predictor in terms of inner products only is instrumental for  mapping it into its kernel-based version. Although the mapping entails the eigenfunctions $\{\phi_i(x)\}$, these are not explicitly present in  \eqref{kernel_trick}, which is given solely  in terms of $k(x,x')$. This is crucial since  $\bmp$ can be infinite dimensional which would render the  method computationally  intractable,  and more importantly the explicit form of $\phi_i(x)$  may not be available. Use of  kernel trick was demonstrated in the context of ridge regression. However, the trick can be used in any vectorial regression or classification method  whose result can be expressed in terms of inner products only. One such example is offered by support vector machines, which find a kernel-based version of the optimal linear classifier in the sense of  minimizing  Vapnik's $\epsilon$-insensitive Hinge loss function, and can be shown  equivalent to the Lasso \cite{G98}.

In a nutshell, the kernel trick provides a means of designing  KBL algorithms, both for nonparametric function estimation [cf. \eqref{representer}], as well as for classification.

\subsection{KBL vis \`a vis Nyquist-Shannon Theorem}
Kernels can be clearly viewed as interpolating bases [cf. \eqref{kernel_expansion}]. This viewpoint can be further  appreciated if one considers the family of bandlimited functions $\Bpi:= \{f\in \mathcal L^2(\calX):\ \int f(x)e^{-i\omega x}dx =0,\ \forall |\omega|>\pi\}$, where  $\mathcal L^2$ denotes the class of square-integrable functions  defined over $\calX=\mathbb R$ (e.g., continuous-time, finite-power signals). The family $\Bpi$ constitutes a linear space. Moreover, any  $f\in\Bpi$ can be generated as the linear combination (span) of $\sinc$ functions; that is,  $f(x)=\sum_{n\in \mathbb Z} f(n)\sinc(x-n)$. This is the cornerstone of signal processing, namely the NST for sampling and reconstruction, but can be viewed also under the lens  of RKHS with $k(x,x')=\sinc(x-x')$ as a reproducing kernel \cite{NS12}. The following  properties (which are proved in the Appendix) elaborate further on this connection.

\noindent\textbf{P1.} The $\sinc$-kernel Gram matrix  $\mathbf K\in\mathbb R^{N\times N}$  satisfies $\mathbf K\succeq \mathbf 0$.\\
\noindent\textbf{P2.} The $\sinc$ kernel decomposes over orthonormal eigenfunctions  $\{\phi_{n}(x)=\sinc(x-n),\ n\in \mathbb Z\}$.\\
\noindent\textbf{P3.} The RKHS norm is \label{prop3}$\|f\|_{\calHx}^2=\int  f^2(x) dx$.

P1 states that  $\sinc(x-x')$  qualifies as a kernel, while  P2 characterizes the eigenfunctions  used in the kernel trick, and P3 shows that the RKHS norm is the restriction of the $\mathcal L^2$ norm to $\Bpi$.

P1-P3  establish that the space of bandlimited functions $\Bpi$ is indeed an RKHS. Any $f\in\Bpi$ can thus be decomposed as a numerable combination of eigenfunctions, where the coefficients and eigenfunctions obey the NST.  Consequently, existence  of eigenfunctions $\{\phi_n(x)\}$ spanning $\Bpi$  is a direct consequence of $\Bpi$ being a RKHS, and does not require the NST unless an explicit form for $\phi_n(x)$ is desired. Finally,  strict adherence to  NST requires  an infinite number of samples  to reconstruct $f\in\Bpi$. Alternatively, the Representer Theorem  fits $f\in\Bpi$ to a finite set of (possibly noisy) samples by regularizing the power of $f$.

\section{Sparse additive nonparametric modeling}

The account of sparse KBL methods begins with  SpAMs and MKL approaches. Both model the function to be learned as  a sparse sum of nonparametric components, and both rely on group Lasso to find it. The additive models considered in this section will naturally lend themselves to  the general model for NBP introduced  in Section IV, and used henceforth.

\subsection{SpAMs for High-Dimensional Models}
Additive function models offer a generalization of linear regression to the nonparametric setup, on the premise of dealing with   \emph{the curse of dimensionality,} which is inherent  to learning  from high dimensional data \cite{HTF09}.

Consider learning a  multivariate function $f:\calX \to \mathbb R$ defined over the Cartesian product  $\calX:=\calX_1\otimes \ldots\otimes\calX_P$  of measurable spaces $\calX_i$. Let $\bx^T:=[x_1,\ldots,x_P]$ denote a point in $\mathcal X$,  $k_i$ the kernel defined over $\calX_i\times \calX_i$,  and $\mathcal H_i$ its associated  RKHS.
 Although $f(\mathbf x)$ can be interpolated from data  via \eqref{representer} after substituting $\mathbf x$ for  $x$, the fidelity of \eqref{kernel_expansion}  is  severely degraded in high dimensions. Indeed, the accuracy of \eqref{kernel_expansion}  depends on  the availability of   nearby points $\bx_n$, where the function is fit to the  (possibly noisy) data $z_n$. But proximity of points $\bx_n$ in high dimensions is challenged by the curse of dimensionality, demanding an excessively large dataset.  For instance, consider positioning  $N$ datapoints randomly in the hypercube $[0,1]^P$,  repeatedly for $P$ growing unbounded and $N$ constant. Then  $\lim_{P\to\infty}\min_{n\neq n'}\mathbf E \|\bx_n-\bx_{n'}\|=1$; that is, the expected distance between any two points is equal to the side of the hypercube \cite{HTF09}.

To overcome this problem,  an additional modeling assumption is well motivated, namely constraining  $f(\bx)$  to the  family of separable functions of the form
\begin{equation}
 f(\bx)=\sum_{i=1}^P c_i(x_i)\label{additive}
 \end{equation}
with $c_i\in\mathcal H_i$
 depending only on the $i$-th entry of  $\bx$, as in e.g., linear regression  models $f_{\textrm{linear}}(\bx):=\sum_{i=1}^P \beta_i x_i$. With $f(\bx)$   separable as in $\eqref{additive}$,  the interpolation task is split into  $P$ one-dimensional problems that are not affected by the curse of dimensionality.

The additive form in \eqref{additive} is also amenable to subsect selection, which yields a  SpAM. As in sparse linear regression, SpAMs involve functions  $f$ in \eqref{additive} that can be expressed using only a few  entries of $\bx$. Those  can be learned using  a variational version of the Lasso  given by \cite{RLLW09}
\begin{equation}\label{nplasso}
\hat f =\arg\min_{f\in\mathcal F_P}\frac{1}{2}\sum_{n=1}^N (z_n-f(\bx_n))^2+\mu\sum_{i=1}^P \|c_i\|_{\mathcal H_i}
\end{equation}
where $\mathcal F_P:=\{f:\calX\to\mathbb R:\ f(\bx)=\sum_{i=1}^Pc_i(x_i)\}$.

With $x_{ni}$ denoting the  $i$th entry of $\mathbf x_n$,  the Representer Theorem \eqref{general_representer} can be applied per component $c_i(x_i)$ in \eqref{nplasso}, yielding  kernel expansions $\hat c_i(x_i)=\sum_{n=1}^N \gamma_{ni} k_i(x_{ni},x_i)$ with scalar coefficients $\{\gamma_{ni},\ i=1,\ldots,P, \ n=1,\ldots,N\}$.
The fact that \eqref{nplasso}  yields a SpAM is demonstrated by substituting these expansions back into \eqref{nplasso} and solving for $\gammai^T:=[\gamma_{i1},\ldots,\gamma_{iN}]$, to obtain
\begin{equation}\label{glasso_gamma}
\{\hat\gammai\}_{i=1}^P =\arg\hspace{-0.1cm}\min_{\{\gammai\}_{i=1}^P}\frac{1}{2}\left\|\mathbf z -\textstyle{\sum_{i=1}^P} \mathbf K_i\gammai \right\|_2^2+\mu\sum_{i=1}^P \|\gammai\|_{\mathbf K_i}
\end{equation}
where  $\mathbf K_i$ is the Gram matrix associated with  kernel $k_i$, and $\|\cdot \|_{\mathbf K_i}$ denotes the weighted $\ell_2$-norm  $\|\bm\gamma_i \|_{\mathbf K_i}:=(\bm\gamma_i^T\mathbf K_i\bm\gamma_i)^{1/2}$.

\subsection{Nonparametric Lasso}
Problem \eqref{glasso_gamma} constitutes a weighted version of the group Lasso formulation for sparse linear regression. Its solution can be found either via  block coordinate descent (BCD) \cite{RLLW09}, or by substituting $\bm \gamma'_i=\mathbf K_i^{1/2} \bm\gamma_i$  and applying the alternating-direction method of multipliers (ADMM) \cite{BMG11}, with convergence guaranteed by its convexity and the  separable structure of the its non-differentiable term \cite{TS09}.
 In any case,  group Lasso regularizes sub-vectors $\gammai$ separately, effecting group-sparsity in the estimates; that is, some of the vectors $\hat\gammai$ in \eqref{glasso_gamma} end up being identically zero. To gain intuition on this, \eqref{glasso_gamma} can be rewritten using the change of variables $\mathbf K_i^{1/2}\gammai=t_i\mathbf u_i$, with $t_i\geq 0$ and $\|\mathbf u_i\|=1$. It will be argued that if $\mu$ exceeds a threshold, then the optimal $t_i$ and thus $\hat\gammai$ will be null. Focusing on the minimization of \eqref{glasso_gamma} w.r.t. a particular sub-vector $\gammai$, as in a BCD algorithm, the substitute variables  $t_i$ and $\mathbf u_i$ should minimize
\begin{equation}\label{glasso_t}
\frac{1}{2}\left\|\mathbf{ z}_i -  \mathbf K_i^{1/2}t_i\mathbf u_i \right\|_2^2+\mu t_i
\end{equation}
where $\mathbf{ z}_i:=\mathbf{ z}-\sum_{j\neq i} \mathbf K_j\bm\gamma_j.$
Minimizing  \eqref{glasso_t} over $t_i$ is a convex univariate problem whose solution lies  either at the border of the constraint,  or, at a stationary point; that is,
\begin{equation}\label{solving_ti}
t_i=\max\left\{0,\frac{\mathbf z_i^T\mathbf K_i^{1/2}\mathbf u_i-\mu}{\mathbf u_i^T \mathbf K_i \mathbf u_i}\right\}.
\end{equation}
The Cauchy-Schwarz inequality implies that   $\mathbf z_i^T\mathbf K_i^{1/2}\mathbf u_i \leq \|\mathbf K_i^{1/2}\mathbf z_i\|$ holds for any $\mathbf u_i$ with $\|\mathbf u_i\|=1$. Hence, it follows from \eqref{solving_ti} that if $\mu\geq  \|\mathbf K_i^{1/2}\mathbf z_i\|$, then $t_i=0$, and thus $\bm \gammai=\mathbf 0$.

The sparsifying effect of \eqref{nplasso} on the additive model \eqref{additive} is now revealed. If $\mu$ is selected large enough, some of the optimal sub-vectors $\hat\gammai$ will be null, and the corresponding functions $\hat c_i(x_i)=\sum_{n=1}^N \hat\gamma_{ni} k(x_{ni},x_i)$ will be identically zero in  \eqref{additive}.  Thus, estimation via  \eqref{nplasso} provides a nonparametric counterpart of Lasso, offering the flexibility of selecting the most informative component-function regressors in the additive model.

The separable structure postulated in \eqref{additive} facilitates subset selection in the nonparametric setup, and mitigates the problem of interpolating scattered data in high dimensions.  However, such a model reduction may render \eqref{additive}  inaccurate, in which case extra components depending on two or more variables can be added, turning \eqref{additive} into the ANOVA model \cite{LZ06}.


%
%
%
\subsection{Multi-Kernel Learning}
Specifying the kernel that ``shapes'' $\calHx$, and thus judiciously determines $\hat f$ in \eqref{representer} is a prerequisite for KBL. Different candidate kernels $k_1,\ldots,k_P$ would  produce different function estimates. Convex combinations  can be also employed in \eqref{representer}, since  elements of the convex hull $\mathcal K :=\{k=\sump a_i k_i, \ a_i\geq 0,\ \sump a_i=1\}$ conserve the defining properties of  kernels.

A data-driven strategy to select ``the best'' $k\in \mathcal K$ is to incorporate the kernel as a variable in \eqref{general_representer}, that is \cite{KY10}
\begin{equation}
\vspace*{-0.2cm}
\hat f=\arg\min_{k\in \mathcal K , f\in \calHx^k} \sumn(z_{n}-f(x_n))^2+\mu \|f\|_{\calHx^k} \label{representerK}
\end{equation}
where the notation $\calHx^k$ emphasizes dependence on $k$.

Then, the following Lemma  brings MKL to the ambit of sparse additive nonparametric models.

\begin{lemma}[\cite{MP05}]\label{lemma1}
Let $\{k_1,\ldots,k_P\}$ be a set of kernels  and $k$ an element of their convex hull $\mathcal K$. Denote by  $\calHi$ and $\calHx^k$  the RKHSs corresponding to $k_i$ and $k$, respectively,   and by $\calHx$ the direct sum $\calHx:= \mathcal H_1\oplus\ldots\oplus\mathcal H_P.$ It then holds that:
\hspace{-1cm}\begin{enumerate} \item[a)]
$\calHx^k=\calHx,\ \forall k\in\mathcal K$; and
\item[b)]  $\forall\  f,\ \inf\{\|f\|_{\calHx^k}:\ k\in \mathcal K\}=\min\{\sump \|c_i\|_{\calHi}:\ f=\sump c_i,\ c_i\in \calHi\}$.
\end{enumerate}
\end{lemma}

According to Lemma \ref{lemma1}, $\calHx$ can replace $\calHx^k$ in \eqref{representerK}, rendering it equivalent to
\begin{align}
\hat f=&\arg\min_{f\in \calHx} \sumn(z_{n}-f(x_n))^2+\mu \sump \|c_i\|_{\calHi} \label{mkl}\\
\nonumber&\sto \{f=\sump c_i,\ c_i\in \calHi, \ \calHx:= \mathcal H_1\oplus\ldots\oplus\mathcal H_P \}.
\end{align}

MKL as in \eqref{mkl} resembles \eqref{nplasso}, differing in that components $c_i(x)$ in \eqref{mkl} depend on the same variable $x$. Taking into account this difference,  \eqref{mkl} is reducible to \eqref{glasso_gamma} and thus solvable via BCD or ADMoM, after  substituting   $k_i(x_n,x)$ for  $k_i(x_{ni},x_i)$.  On the other hand, a more general case of MKL is presented in \cite{MP05}, where $\mathcal K$ is the convex hull of an infinite and possibly uncountable family of kernels.

 An example of   MKL applied to wireless communications is offered   in Section \ref{sec:applications}, where two different kernels are employed for estimating path-loss and shadowing propagation effects in a cognitive radio sensing paradigm.

In the ensuing section, basis functions  depending on a second variable $y$  will be incorporated to broaden the scope of the additive models just described.
\section{Nonparametric basis pursuit}
 Consider  function $f:\calX\times\calY\to\mathbb R$ over the Cartesian product of spaces $\calX$ and $\calY$ with associated RKHSs $\calHx$ and $\calHy$, respectively. Let $f$ abide to the bilinear  expansion form
\begin{equation}
f(x,y)=\sum_{i=1}^P c_i(x)b_i(y)\label{bilinear}
\end{equation}
where  $b_i:\calY\to\mathbb R$ can be viewed as bases, and  $c_i:\calX\to\mathbb R$ as expansion coefficient functions.
Given a finite number of training data,  learning $\{c_i,b_i\}$ under sparsity constraints constitutes the goal of the NBP approaches developed in the following sections.

The first method for sparse KBL of $f$ in \eqref{bilinear} is related to a \emph{nonparametric} counterpart of basis pursuit, with the goal of fitting the function $f(x,y)$  to data,  where  $\{b_i\}$ are prescribed and $\{c_i\}$s are to be learned.
 The designer's degree of confidence on the modeling assumptions is key to deciding whether $\{b_i\}$s should be prescribed or learned from data. If the prescribed $\{b_i\}$s are unreliable, model \eqref{bilinear} will be inaccurate and the  performance of KBL will suffer. But neglecting the prior knowledge conveyed by $\{b_i\}$s may be also damaging. Parametric basis pursuit \cite{CDS98} hints toward addressing this tradeoff  by offering a compromising alternative.

 A functional dependence $z=f(y)+e$ between input $y$ and output $z$ is modeled  in \cite{CDS98} with an overcomplete set of bases $\{b_i(y)\}$ (a.k.a. regressors) as
\begin{align}
z=\sump c_i b_i(y) + e,\ ~~~e\sim\mathcal N(0,\sigma^2).
\end{align}
Certainly, leveraging an overcomplete set of bases $\{b_i(y)\}$ can accommodate uncertainty. Practical merits of basis pursuit however, hinge on  its capability to learn the few $\{b_i\}$s that ``best''  explain the given data.

 The crux of NBP on the other hand, is to fit $f(x,y)$ with a basis expansion over the  $y$ domain, but learn its dependence on $x$  through nonparametric means. Model \eqref{bilinear} comes handy for this purpose, when $\{b_i(y)\}_{i=1}^P$ is a generally overcomplete collection of prescribed bases.

  With $\{b_i(y)\}_{i=1}^P$ known, $\{c_i(x)\}_{i=1}^P$ need to be estimated, and a kernel-based strategy can be adopted to this end.  Accordingly, the optimal function $\hat f(x,y)$ is searched over the family $\mathcal F_b:=\{f(x,y)=\sum_{i=1}^Pc_i(x)b_i(y)\}$, which constitutes the feasible set for the NBP-tailored nonparametric Lasso [cf. \eqref{nplasso}]
\begin{equation}\label{nbp}
\hat f =\arg\min_{f\in \mathcal F_b}\sum_{n=1}^N (z_n-f(x_n,y_n))^2+\mu\sum_{i=1}^P \|c_i\|_{\calHx}.
\end{equation}

The Representer Theorem in its general form  \eqref{general_representer} can be applied recursively to minimize \eqref{nbp} w.r.t. each $c_i(x)$ at a time, rendering $\hat f$ expressible in terms of the kernel expansion  as $\hat f(x,y)=\sum_{i=1}^P\sum_{n=1}^N\gamma_{in}k(x_n,x)b_i(y)$, where coefficients $\gammai^T:=[\gamma_{i1},\ldots,\gamma_{iN}]$ are learned from data  $\mathbf z^T:=[z_1,\ldots,z_N]$  via group Lasso [cf. \eqref{glasso_gamma}]
\begin{equation}\label{group_lasso_nbp}
\min_{\{ \gammai\in \mathbb R^N \}_{i=1}^P}\left\|\mathbf z-\textstyle{\sump} \mathbf K_i \gammai\right\|^2+\mu\sum_{i=1}^P \|\gammai\|_{\mathbf K}
\end{equation}
with $\mathbf K_i:=\textrm{{Diag}}[b_i(y_1),\ldots,b_i(y_N)]\mathbf K$.

As it was argued in Section III, group Lasso in \eqref{group_lasso_nbp} effects group-sparsity in the subvectors $\{\gammai\}_{i=1}^{P}$. This property  inherited by \eqref{nbp} is the capability of selecting bases in the nonparametric setup. Indeed, by zeroing  $\gammai$  the corresponding coefficient function $c_i(x)=\sumn \gamma_{in}k(x_n,x)$ is driven to zero, and correspondingly  $b_i(y)$ drops from the expansion \eqref{bilinear}.


\noindent\textbf{Remark 2.} A single kernel  $k_{\calX}$ and associated RKHS $\calHx$ can be used for all components $c_i(x)$ in \eqref{nbp}, since the summands in \eqref{bilinear} are differentiated through the bases. Specifically, for a common $\mathbf K$,  a different $b_i(y)$ per coefficient  $c_i(x)$, yields a distinct diagonal matrix $\textrm{{Diag}}[b_i(y_1),\ldots,b_i(y_N)]$, defining an individual $\bK_i$ in \eqref{group_lasso_nbp} that renders  vector $\gammai$  identifiable. This is a particular characteristic of \eqref{nbp}, in contrast with \eqref{nplasso} and Lemma \ref{lemma1} which are designed for, and require, multiple kernels.

\noindent\textbf{Remark 3.} The different sparse kernel-based approaches presented so far, namely SpAMs, MKL, and NBP, should not be viewed as competing but rather as complementary choices. Multiple kernels can be used in basis pursuit, and a separable model for $c_i(x)$ may be due in high dimensions.  An NBP-MKL hybrid applied to spectrum cartography   illustrates this point in Section \ref{sec:applications}, where bases are utilized for the frequency domain $\mathcal Y$.

\section{Blind NBP for matrix and tensor completion}
A kernel-based matrix completion scheme will be developed in this section using  a \emph{blind} version of NBP, in which bases $\{b_i\}$ will not be prescribed, but they will be learned together with coefficient functions $\{c_i\}$.  The matrix completion task entails imputation of missing entries of a data matrix $\mathbf Z\in\mathbb R^{M\times N}$. Entries of an index matrix $\mathbf W\in\{0,1\}^{M\times N}$ specify whether datum $z_{mn}$ is available ($w_{mn}=1$), or missing  ($w_{mn}=0$). Low rank of $\bZ$  is a popular attribute  that relates missing with available data, thus granting feasibility to the  imputation task.  Low-rank matrix imputation  is achieved by solving
\begin{align}\label{low-rank}
\hat{\mathbf Z}=\arg\min_{\mathbf A\in \mathbb R^{M\times N}}&\frac{1}{2}\|(\mathbf Z-\mathbf A)\odot\mathbf W\|_F^2
\textrm{ s. to rank}(\mathbf A)\leq P
\end{align}
where $\odot$ stands for the Hadamard (element-wise) product.
The low-rank constraint corresponds to an upperbound on the number of nonzero singular values of matrix $\mathbf A$, as given by its $\ell_0$-norm. Specifically, if  $\mathbf s^T:=[s_1,\ldots,s_{\min\{M,N\}}]$ denotes vector of singular values of $\mathbf A$, and the cardinality $|\{s_i\neq 0, \ i=1,\ldots, \min\{M,N\}\}|:=\|\mathbf s\|_0$ defines its $\ell_0$-norm, then the ball of radius $P$, namely $\|\mathbf s\|_0\leq P$, can replace $\textrm{rank}(\mathbf A)\leq P$ in \eqref{low-rank}. The feasible set $\|\mathbf s\|_0\leq P$ is not convex because $\|\mathbf s\|_0$ is not a proper norm (it lacks linearity), and solving \eqref{low-rank}  requires a combinatorial search for the  nonzero entries of $\mathbf s$. A convex relaxation is thus well motivated. If the $\ell_0$-norm is surrogated  by the $\ell_1$-norm,  the corresponding ball $\|\mathbf s\|_1\leq P$ becomes the convex hull of the original feasible set. As the singular values of $\mathbf A$ are non-negative by definition, it follows that $\|\mathbf s\|_1=\sum_{i=1}^{\min\{M,N\}}s_i$. Since the sum of  singular values equals the dual norm of the $\ell_2$-norm of $\mathbf A$ \cite[p.637]{BV04},  $\|\mathbf s\|_1$ defines a norm over the matrix $\mathbf A$ itself, namely the nuclear norm of $\mathbf A$,   denoted by $\|\mathbf A\|_*$.

Upon substituting $\|\mathbf A\|_*$ for the rank, \eqref{low-rank} is further   transformed to its Lagrangian form by placing the constraint in the objective as a regularization term, i.e.,
\begin{align}\label{nuclear norm}
\hat{\mathbf Z}=\arg\min_{\mathbf A\in \mathbb R^{M\times N}}&\frac{1}{2}\|(\mathbf Z-\mathbf A)\odot\mathbf W\|_F^2+\mu\|\mathbf A\|_*.
\end{align}

The next step towards kernel-based matrix completion relies on an alternative definition of $\|\mathbf A\|_*$. Consider bilinear  factorizations of matrix $\mathbf A=\bC\bB^T$ with  $\mathbf B\in\mathbb R^{N\times P}$ and $\mathbf C\in\mathbb R^{M\times P}$, in which the constraint $\textrm{rank}(\mathbf A)\leq P$ is implicit. The nuclear norm of $\mathbf A$ can be redefined as (see e.g., \cite{MMG12})
\begin{equation}\label{infimum}
\|\mathbf A\|_*=\inf_{ \mathbf A=\mathbf C\mathbf B^T }{\frac{1}{2}(\|\mathbf B\|_F^2+\|\mathbf C\|_F^2)}.
\end{equation}
Result \eqref{infimum} states that the infimum is attained by the singular value decomposition of $\mathbf A$. Specifically, if  $\mathbf A  =\mathbf U \mathbf \Sigma\mathbf V^T $ with $\mathbf U$ and $\mathbf V$ unitary and  $\mathbf \Sigma:=\textrm{diag}(\mathbf s),$ and if $\mathbf B$ and $\mathbf C$ are selected as $\mathbf B=\mathbf V\mathbf \Sigma^{1/2}$, and $\mathbf C=\mathbf U\mathbf \Sigma^{1/2}$, then $\frac{1}{2}(\|\mathbf B\|_F^2+\|\mathbf C\|_F^2)=\sum_{i=1}^{P} s_i = \|\mathbf A\|_*.$
Given \eqref{infimum}, it is possible to rewrite \eqref{nuclear norm} as
\begin{align}\label{nuclear norm_BC}
\hat{\mathbf Z}=\arg\min_{\mathbf A=\bC\bB^T}&\frac{1}{2}\|(\mathbf Z-\mathbf A)\odot\mathbf W\|_F^2+\frac{\mu}{2}(\|\mathbf B\|_F^2+\|\mathbf C\|_F^2).
\end{align}
A formal proof of the equivalence between \eqref{nuclear norm} and \eqref{nuclear norm_BC} can be found in \cite{MMG12}.

Matrix completion in its factorized form \eqref{nuclear norm_BC} can be reformulated in terms of \eqref{bilinear} and RKHSs. Following  \cite{ABEV09}, define spaces $\calX:=\{1,\ldots,M\}$ and $\calY:=\{1,\ldots,N\}$ with associated kernels $k_{\calX}(m,m')$ and  $k_{\calY}(n,n'),$ respectively. Let $f(m,n)$ represent the  $(m,n)$-th entry of the approximant matrix $\mathbf A$ in \eqref{nuclear norm_BC}, and $P$ a prescribed overestimate of its rank. Consider estimating $f:\calX\times\calY\to \mathbb R$ in \eqref{bilinear} over the family $\mathcal F :=\{f(m,n)=\sump c_i(n)b_i(m),\ c_i\in \calHx,\ b_i\in\calHy\}$  via
\begin{align}
\hat f=\arg\min_{f\in\mathcal F} \frac{1}{2}&\sum_{m=1}^M\sumn w_{mn}(z_{mn}-f(m,n))^2\nonumber\\
&+\frac{\mu}{2} \sump \left(\|c_i\|^2_{\calHx}+\|b_i\|^2_{\calHy}\right) \label{nuclear_norm_delta}.
\end{align}

If both kernels are selected as Kronecker delta functions, then  \eqref{nuclear_norm_delta} coincides with \eqref{nuclear norm_BC}. This equivalence is stated in the following lemma.


\begin{lemma}\label{lemma2}
Consider spaces $\calX:=\{1,\ldots,M\},$   $\calY:=\{1,\ldots,N\}$ and kernels $k_{\calX}(m,m'):=\delta(m-m')$ and $k_{\calY}(n,n'):=\delta(n-n')$ over the product spaces $\calX\times\calX$ and $\calY\times\calY$, respectively. Define functions $f:\calX\times\calY\to \mathbb R$, $c_i:\calX\to \mathbb R$, and $b_i:\calY\to \mathbb R, \ i=1,\ldots, P$, and matrices $\mathbf A \in \mathbb R^{M\times N}$, $\mathbf B \in \mathbb R^{N\times P},$ and $\mathbf C \in \mathbb R^{M\times P}.$
It holds that:
\begin{enumerate}
\item[a)]  RKHS $\calHx$ ($\calHy$) of functions over $\calX$ (correspondingly $\calY$), associated with $k_{\calX}$ ($k_{\calY}$) reduce to $\calHx=\mathbb R^M$ ($\calHy=\mathbb R^N$).
    \item[b)] Problems \eqref{nuclear_norm_delta},  \eqref{nuclear norm_BC}, and  \eqref{nuclear norm} are equivalent upon identifying $f(m,n)=A_{mn}$, $b_i(n)=B_{ni}$, and $c_i(m)=C_{mi}.$
 \end{enumerate}
\end{lemma}

According to Lemma \ref{lemma2}, the intricacy of rewriting \eqref{nuclear norm} as in \eqref{nuclear_norm_delta} does not introduce any benefit when the kernel is selected as the Kronecker delta. But as it will be argued next, the equivalence between these two estimators generalizes nicely the matrix completion problem to sparse KBL of missing data with arbitrary kernels.

The separable structure of the regularization term in \eqref{nuclear_norm_delta} enables a finite dimensional representation of  functions
 \begin{align}
 \nonumber   \hat c_i(m)&=\sum_{m'=1}^M \gamma_{im'}k_{\calX}(m',m),\   m=1,\ldots,M, \\
    \hat b_i(n)&=\sum_{n'=1}^N \beta_{in'}\ky(n',n),\   n=1,\ldots,N. \label{c_and_b}
\end{align}

Optimal scalars $\{\gamma_{im}\}$ and $\{\beta_{in}\}$ are obtained by substituting \eqref{c_and_b} into \eqref{nuclear_norm_delta}, and solving
\begin{align}
\min_{\substack{\tbC\in \mathbb R^{M\times P}\\\tbB\in \mathbb R^{N\times P}}}&\frac{1}{2}\|(\bZ-\Kx\tbC\tbB^T\Ky^T)\odot\bW\|_F^2\nonumber\\&+\frac{\mu}{2} \left[\tr(\tbC^T\Kx\tbC)+\tr(\tbB^T\Ky\tbB)\right]\label{coefs_nmc}
\end{align}
where matrix  $\tbC$ ($\tbB$) is formed  with entries $\gamma_{mi}$ ($\beta_{ni}$).

A Bayesian approach to  kernel-based matrix completion  is given next, followed by an algorithm to solve for $\tbB$ and $\tbC$.

\subsection{Bayesian Low-Rank Imputation and Prediction}
To recast  \eqref{nuclear_norm_delta} in a Bayesian framework, suppose that the available entries of $\bZ$ obey the additive white Gaussian noise  (AWGN) model
$\bZ=\mathbf A+\mathbf E,$
 with $\mathbf E$ having   entries  independent identically distributed  (i.i.d.) according to the zero-mean Gaussian distribution  $\mathcal N(0,\sigma^2)$.

Matrix $\mathbf  A$ is factorized  as $\mathbf  A=\bC\bB^T$ without loss of generality (w.l.o.g.).  Then, a Gaussian prior is assumed for each of the columns $\mathbf b_i$ and $\mathbf c_i$ of  $\bB$ and  $\bC$, respectively,
\begin{align}\label{prior}
\mathbf b_i &\sim \mathcal N(\mathbf 0,\mathbf R_B),\
\mathbf c_i \sim \mathcal N(\mathbf 0,\mathbf R_C)
 \end{align}
 independent across $i,$ and with  $\tr(\mathbf R_B)=\tr(\mathbf R_C)$. Invariance across $i$ is justifiable, since columns are a priori interchangeable, while $\tr(\mathbf R_B)=\tr(\mathbf R_C)$ is introduced w.l.o.g. to remove the scalar ambiguity in $\mathbf A=\bC\bB^T$.

Under the AWGN model, and with priors \eqref{prior}, the maximum a posteriori (MAP) estimator of $\mathbf  A$ given $\mathbf  Z$ at the entries indexed by $\bW$ takes the form [cf. \eqref{coefs_nmc}]
\begin{align}
\min_{\substack{\bC\in \mathbb R^{M\times P}\\\bB\in \mathbb R^{N\times P}}}&\frac{1}{2}\|(\bZ-\bC\bB^T)\odot\bW\|_F^2\nonumber\\&+\frac{\sigma^2}{2} \left[\tr(\bC^T\mathbf R_C^{-1}\bC)+\tr(\bB^T\mathbf R_B^{-1}\bB)\right].\label{bayesian_nmc}
\end{align}

With $\mathbf R_C=\Kx$ and $\mathbf R_B=\Ky$, and substituting $\bB:=\Ky\tbB$ and $\bC:=\Kx\tbC$, the MAP estimator that solves \eqref{bayesian_nmc} coincides with the estimator solving \eqref{coefs_nmc} for the coefficients of kernel-based matrix completion, provided that covariance and Gram matrices coincide. From this Bayesian perspective, the KBL matrix completion method \eqref{nuclear_norm_delta}  provides a generalization of \eqref{nuclear norm}, which can accommodate a priori knowledge in the form of correlation across rows and columns of the incomplete  $\bZ$.

With prescribed correlation matrices $\mathbf R_B$ and $\mathbf R_C$, \eqref{nuclear_norm_delta} can even perform smoothing and prediction. Indeed, if a column (or row) of $\bZ$ is completely missing, \eqref{nuclear_norm_delta} can still find an estimate $\mathbf{ \hat Z}$ relying on the covariance between the missing and available columns. This feature is not available with \eqref{nuclear norm}, since the latter relies only on rank-induced colinearities, so it cannot reconstruct a missing column. The prediction capability is useful for instance in collaborative filtering \cite{ABEV09}, where a group of users rates  a collection of items, to enable inference  of new-user preferences or items entering the system.
Additionally, the Bayesian reformulation \eqref{bayesian_nmc} provides an explicit interpretation for  the regularization parameter $\mu=\sigma^2$  as the variance of the model error, which  can thus  be obtained from training data.
The kernel-based matrix completion method \eqref{bayesian_nmc} is summarized in Algorithm \ref{KML-table}, which solves \eqref{bayesian_nmc} upon identifying $\mathbf R_C=\Kx$, $\mathbf R_B=\Ky$, and $\sigma^2=\mu$, and solves  \eqref{coefs_nmc} after changing variables $\bB:=\Ky\tbB$ and $\bC:=\Kx\tbC$ (compare \eqref{coefs_nmc} with lines 13-14 in Algorithm \ref{KML-table}).
 \begin{algorithm}[t]
\caption{: Kernel Matrix Completion (KMC)} \small{
 \begin{algorithmic}[1]
\State Initialize $\bB$ and $\bC$ randomly.%
\State Set the identity matrix $\mathbf I_P$,  with dimensions $P\times P$, and columns $\mathbf e_i, \ i=1,\ldots,P$
      \While {$|\textrm{cost}-\textrm{cost\_old}|<\epsilon$ }
             \For{$i=1,\ldots,P$}
            \State Set $\mathbf{Z}_{i}:=\bZ-\bC(\mathbf I_P-\mathbf e_i\mathbf e_i^T)\bB^T$
             \State Compute  $\mathbf H_i:=\textrm{Diag}[\bW(\bB\mathbf e_i\odot\bB\mathbf e_i)]+\mu \mathbf K_{\mathcal Y}^{-1}$
             \State Update column $\mathbf c_i=\mathbf H_i^{-1}(\bW\odot \mathbf{ Z}_i)\bB \mathbf e_i$
                                                             \EndFor

                              \For {$i=1$,$\ldots,P$}
            \State Set $\mathbf{Z}_{i}:=\bZ-\bC(\mathbf I_P-\mathbf e_i\mathbf e_i^T)\bB^T$
            \State Compute $\mathbf{\bar H}_i:=\textrm{Diag}[\bW^T(\bC\mathbf e_i\odot\bC\mathbf e_i)]+\mu \mathbf K_{\mathcal X}^{-1}$
             \State Update column $\mathbf b_i=\mathbf{\bar H}_i^{-1}(\bW^T\odot \mathbf{ Z}_i^T)\bC \mathbf e_i$
                \EndFor
  \State  Recalculate cost $=\frac{1}{2}\|(\bZ-\bC\bB^T)\odot\bW\|_F^2$\\$\hspace{2.5cm}+\frac{\mu}{2} \left[\tr(\bC^T\Kx^{-1}\bC)+\tr(\bB^T\Ky^{-1}\bB)\right]$
         \EndWhile\\
\Return $\tbB=\Ky^{-1}\bB$, $\tbC=\Kx^{-1}\bC$,  and $\mathbf {\hat Z}=\bC\bB^T$
 \end{algorithmic}}
\label{KML-table}
 \end{algorithm}

Detailed derivations of the updates in Algorithm \ref{KML-table} are provided in the Appendix. For a high-level description,   the columns of  $\bB$ and $\bC$ are updated  cyclically,  solving \eqref{bayesian_nmc} via BCD iterations. This procedure converges to a stationary point of \eqref{bayesian_nmc}, which  in principle does not guarantee global optimality. Opportunely, it can be established that local minima
 of \eqref{bayesian_nmc} are global minima, by transforming \eqref{bayesian_nmc} into a convex problem through the same change of variables proposed in \cite{MMG12} for the   analysis of \eqref{nuclear norm_BC}.
This observation implies that Algorithm \ref{KML-table} yields the global optimum of   \eqref{coefs_nmc}, and  thus \eqref{nuclear_norm_delta}.

The kernel-based matrix completion method here offers an alternative to   \cite{ABEV09}, where the low-rank constraint is introduced indirectly through the kernel trick. Furthermore, bypassing the nuclear norm and using \eqref{infimum} instead, renders  \eqref{nuclear_norm_delta}  generalizable to tensor imputation \cite{BMG12}.

\section{Kernel-based dictionary learning}
\begin{figure}
  \begin{center}
  \includegraphics[width=\linewidth]{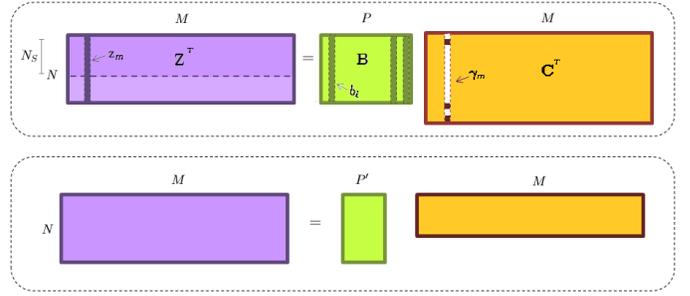}\\
  \caption{Comparison between KDL and NBP; (top) dictionary $\bB$ and sparse coefficients $\bm\gamma_m$ for KDL, where $MN_S$ equations are sufficient to recover $\bC$; (bottom) low-rank structure $\mathbf A=\bC\bB^T$ presumed in KMC. }\label{fig:KDL}
\end{center}
\vspace{-0.6cm}
\end{figure}

Basis pursuit approaches advocate an overcomplete set of bases to cope with model uncertainty, thus learning from data  the most concise subset of bases that represents the signal of interest. But the extensive set of candidate bases (a.k.a. dictionary)  still needs to be prescribed. The next step towards  model-agnostic KBL is to learn the dictionary from data, along with the sparse regression coefficients.
Under the sparse linear model
\begin{align}\label{dictionary_model}
\mathbf z_m = \bB \bm \gamma_m+\mathbf e_m,\ m=1,\ldots,M
\end{align}
with dictionary of bases $\bB\in\mathbb R^{N\times P},$ and vector of coefficients $\bm \gamma_m\in\mathbb R^P$,  the goal of dictionary learning is to obtain  $\bB$ and $\bC :=[\bm\gamma_1,\ldots,\bm\gamma_M]^T$   from data $\bZ:=[\mathbf z_1,\ldots,\bz_M]^T$. A swift count of equations and unknowns yields $NP+MP$ scalar variables to be learned  from $MN$ data (see Fig. \ref{fig:KDL}). This goal is not plausible for an overcomplete design ($P>N$) unless  sparsity  of $\{\bm\gamma_m\}_{m=1}^M$ is exploited.  Under proper conditions, it is  possible to recover a sparse $\bm\gamma_m$  containing at most $S$ nonzero entries from a reduced number $N_s:=\theta S \log P\leq N$ of equations \cite{CT05}, where $\theta$ is a proportionality constant. Hence, the number of equations needed to specify $\bC$ reduces to $MN_s$, as represented by the darkened region of $\bZ^T$ in Fig. \ref{fig:KDL}. With $N_s<N$, it is then possible and crucial to collect a sufficiently large number $M$ of data vectors in order to ensure that $MN\geq NP+MN_S$,  thus  accommodating the additional $NP$ equations needed to determine  $\bB$, and enable learning of the dictionary.

Having collected  sufficient training data, one possible approach  to find $\bB$ and $\bC$ is to fit the data via the LS cost $\|\bZ-\bC\bB^T\|_F^2$ regularized by the $\ell_1$-norm of $\bC$ in order to effect sparsity in the coefficients \cite{KMRELS03}. This dictionary leaning approach can be recast into the form of blind NBP \eqref{nuclear_norm_delta} by introducing  the additional regularizing term $\lambda\sump \|c_i\|_1$, with $\|c_i\|_1:=\sum_{m=1}^M |c_i(m)|$. The new regularizer on functions $c_i:\calX\to\mathbb R$ depends on their values at the measurement points $m$ only, and can be absorbed in the loss part of \eqref{general_representer}. Thus, the  optimal $\{c_i\}$ and $\{b_i\}$ conserve their finite expansion representations  dictated by the Representer Theorem. Coefficients $\{\gamma_{mp},\beta_{np}\}$ must be adapted according to the new cost, and \eqref{bayesian_nmc} becomes
\begin{align}
\label{bayesian_DL}\min_{\substack{\bC\in \mathbb R^{M\times P}\\\bB\in \mathbb R^{N\times P}}}&\frac{1}{2}\|(\bZ-\bC\bB^T)\odot\bW\|_F^2+\lambda\|\bC\|_1\\\nonumber&+\frac{\sigma^2}{2} \left[\tr(\bB^T\mathbf R_B^{-1}\bB)+\tr(\bC^T\mathbf R_C^{-1}\bC)\right].
\end{align}

\noindent\textbf{Remark 4.}
Kernel-based dictionary learning (KDL) via  \eqref{bayesian_DL} inherits two attractive properties of kernel matrix completion (KMC), that is  blind NBP, namely  its flexibility to introduce a priori information through  $\mathbf R_B$ and $\mathbf R_C$, as well as the capability to cope with missing data. While both KDL and KMC estimate  bases $\{b_i\}$ and coefficients $\{c_i\}$  jointly, their   difference lies in the size of the dictionary. As in principal component analysis, KMC presumes a low-rank model for the approximant $\mathbf A=\bC\bB^T$,  compressing signals $\{\mathbf z_m\}$ with $P'<M$ components (Fig. \ref{fig:KDL} (bottom)). Low rank of $\mathbf A$ is not required by the  dictionary learning approach, where signals $\{\mathbf z_m\}$ are spanned by $P\geq M$ dictionary atoms $\{b_i\}$ (Fig. \ref{fig:KDL} (top)), provided that each $\mathbf z_m$ is composed by a few atoms only.

Algorithm \ref{KML-table} can be modified to solve \eqref{bayesian_DL} by replacing the update for column $\mathbf c_i$ in line 7 with the Lasso estimate
\begin{equation}
\mathbf c_i:=\arg\min_{\mathbf c\in \mathbb R^M} \frac{1}{2}\mathbf c^T \mathbf H_i \mathbf c+\mathbf c^T(\bW\odot \mathbf{ Z}_i)\bB \mathbf e_i +\lambda\|\mathbf c\|_1.
\end{equation}

The Bayesian interpretation of \eqref{bayesian_DL} brings KDL close to \cite{XZC12}, where a Bernoulli-Gaussian model for   $\bC$ accounts for its sparsity, and a Beta distribution is introduced for learning the distribution of $\bC$ through hyperparameters. Although \cite{XZC12} assumes independent Gaussian variables across ``time'' samples  in the underlying model for $\bC$,  generalization to correlated variables is straightforward.  Bernoulli parameters controlling the sparsity of $c_{mp}$ are assumed invariant across $m$ in \cite{XZC12}, which amounts to stationarity over $c_{mp}$.

Sparse learning of temporally correlated data is studied also in \cite{ZR11}, although the time-invariant model for the support of $\mathbf c_m$ does not lend itself   to dictionary learning.

Although dictionary learning can indeed be viewed as a blind counterpart of  compressive  sampling, its capability of recovering $\bB$ and  $\bC$ from data is typically illustrated by examples rather than  theoretical guarantees. Recent efforts on establishing identifiability and  local optimality of dictionary learning can be found in \cite{GW11} and \cite{GS12}. A related KDL strategy has been proposed in \cite{SNPC12},   where data and dictionary atoms are organized in classes, and the regularized learning criterion  is designed to promote cohesion of atoms within a class.

\section{Applications}\label{sec:applications}
\subsection{Spectrum cartography via NBP and MKL}
\begin{figure}[ht]
\begin{minipage}[b]{0.45\linewidth}
\centering
\includegraphics[width=4.8cm]{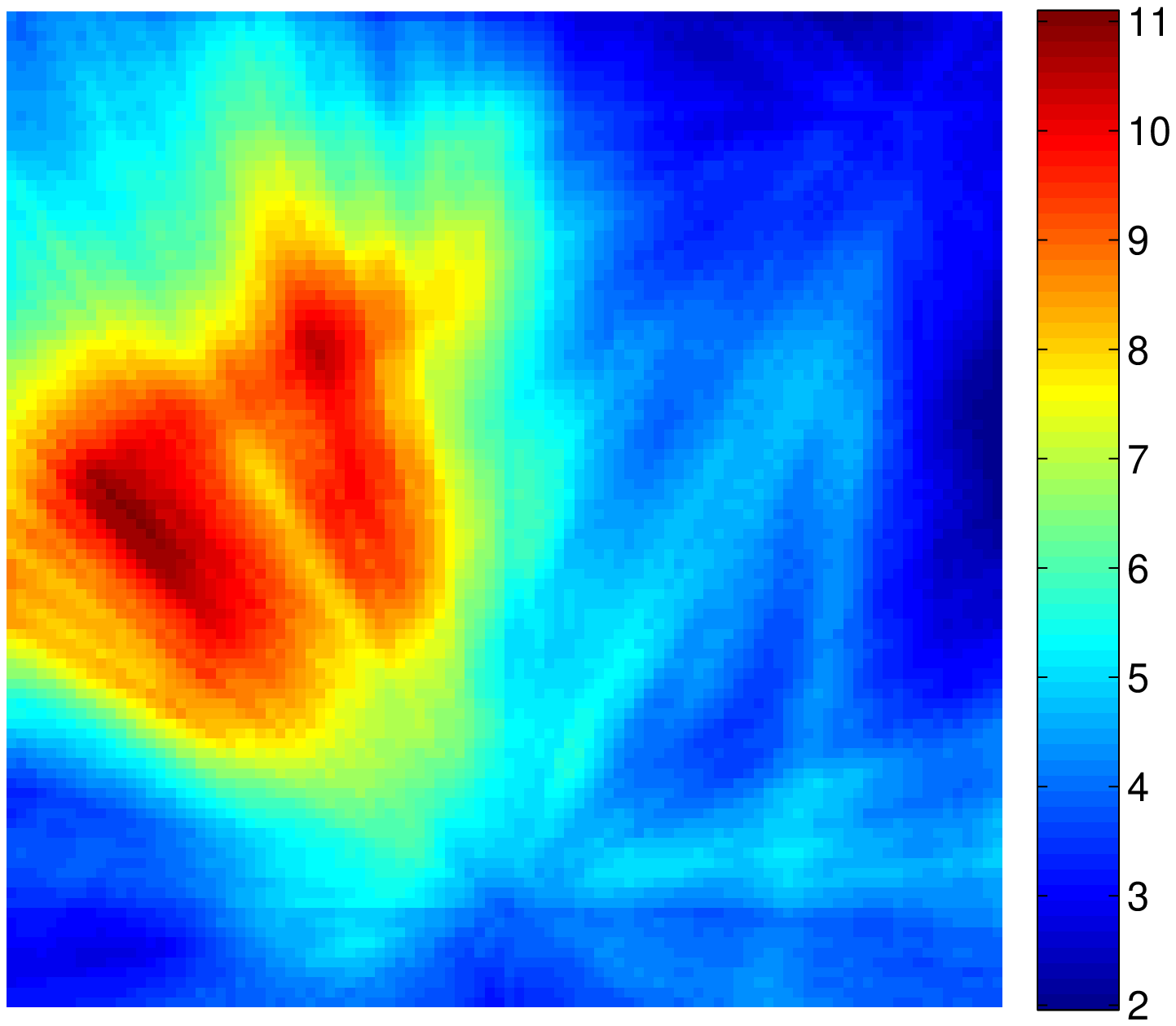}
\caption{ Aggregate power distribution across space.}
\label{fig:power_map}
\end{minipage}
\hspace{0.5cm}
\begin{minipage}[b]{0.45\linewidth}
\centering
\includegraphics[width=4.2cm]{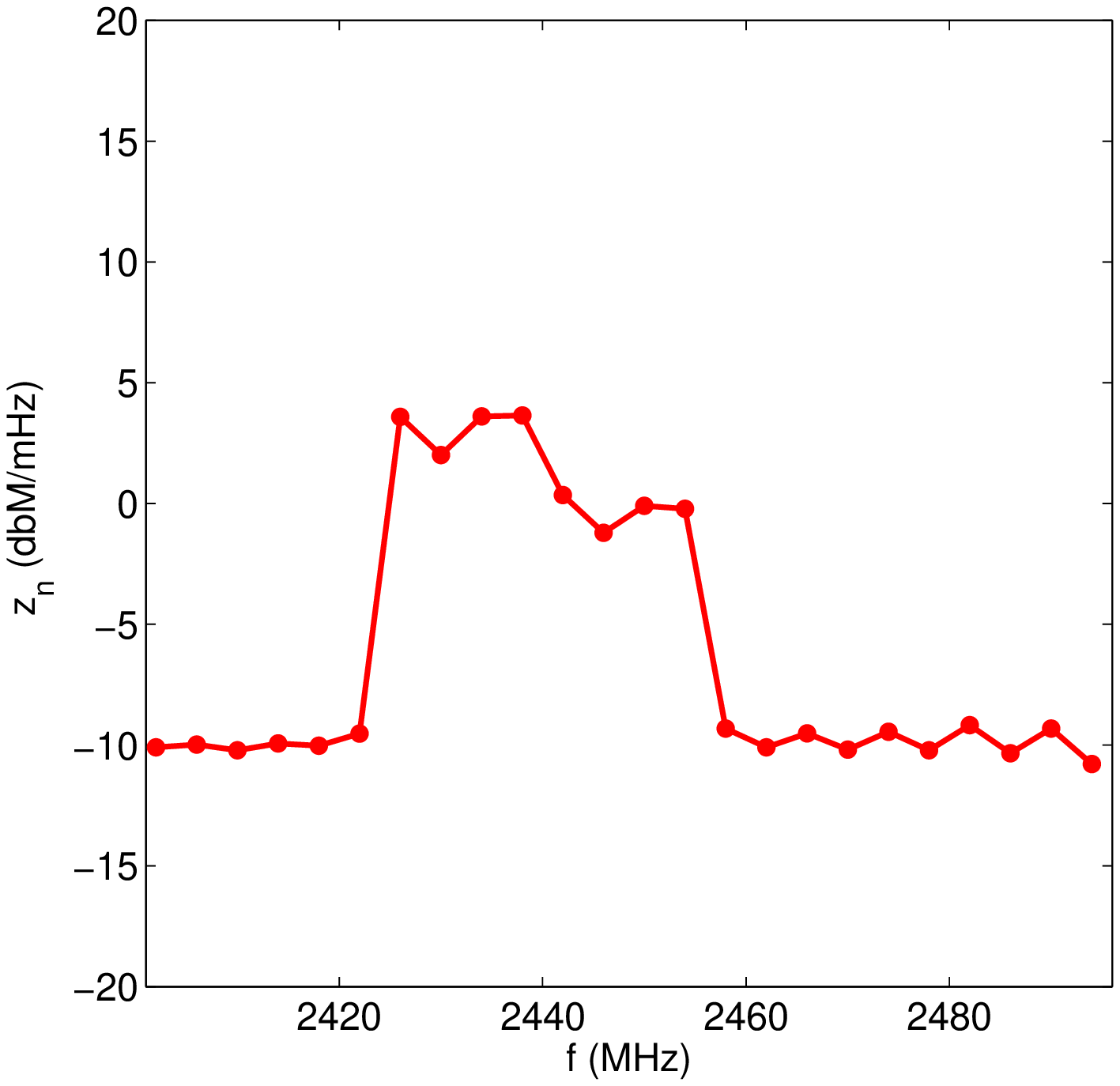}
\caption{ PSD measurements at a representative location $x_n$.}
\label{fig:simulated_data}
\end{minipage}
\end{figure}

Consider the setup in \cite{BMG11} with $N_c=100$ radios distributed over an area $\calX$ of $100\times 100\textrm{m}^2$ to measure the ambient RF power spectral density (PSD) at $N_f=24$ frequencies equally spaced in the band from $2,400$MHz to $2,496$MHz, as specified by IEEE 802.11 wireless LAN standard \cite{IEEE802}. The radios collaborate by sharing their $N=N_cN_f$ measurements with the goal of obtaining a map of the PSD across space and frequency, while specifying at the same time which of the  $P=14$ frequency sub-bands are occupied.  The wireless propagation is simulated according to the pathloss model affected by shadowing  described in \cite{AP09}, with parameters $n_p=3$,  $\Delta_0=60$m,  $\delta=25$m , $\sigma_X^2=25dB$, and  with AWGN variance $\sigma_n^2=-10dB$.
 Fig. \ref{fig:power_map}  depicts the distribution of power across space generated by two sources transmitting over bands $i=5$ and $i=8$ with center frequencies  $2,432$MHz and $2,447$MHz, respectively.  Fig. \ref{fig:simulated_data}  shows the PSD as seen by a representative radio located at the center of $\calX$.

Model \eqref{bilinear} is adopted  for collaborative PSD sensing, with $x$ and $y$ representing the spatial and frequency variables, respectively.  Bases $\{b_i\}$ are prescribed as  Hann-windowed pulses in accordance with \cite{IEEE802}, and the distribution  of power across space per sub-band is given by $\{c_i(x)\}$ after interpolating the measurements obtained by the radios via \eqref{nbp}.
Two exponential kernels $k_r(x,x')=\exp(-\|x-x'\|^2/\theta_r^2),\ r=1,2$ with $\theta_1=10$m and $\theta_2=20$m are selected, and convex combinations of the two are considered as candidate interpolators $k(x,x')$. This MKL strategy is intended for capturing two different levels of resolution as produced by pathloss and shadowing. Correspondingly, each $c_i(x)$ is decomposed into  two functions $c_{i1}(x)$ and $c_{i2}(x)$  which are regularized separately  in $\eqref{nbp}$.

\begin{figure}[htb]
\centering
\includegraphics[width=3cm]{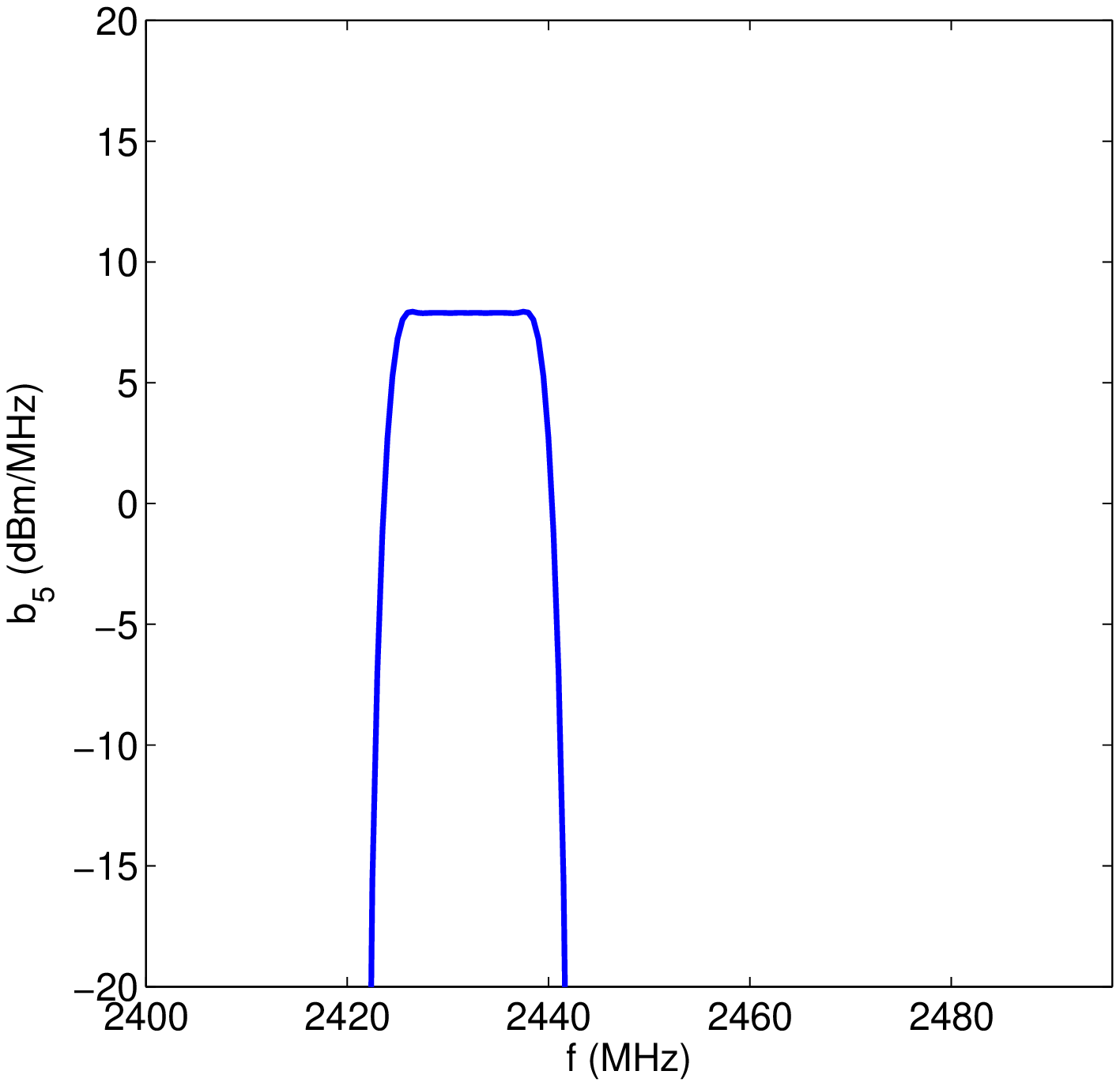}
\includegraphics[width=3cm]{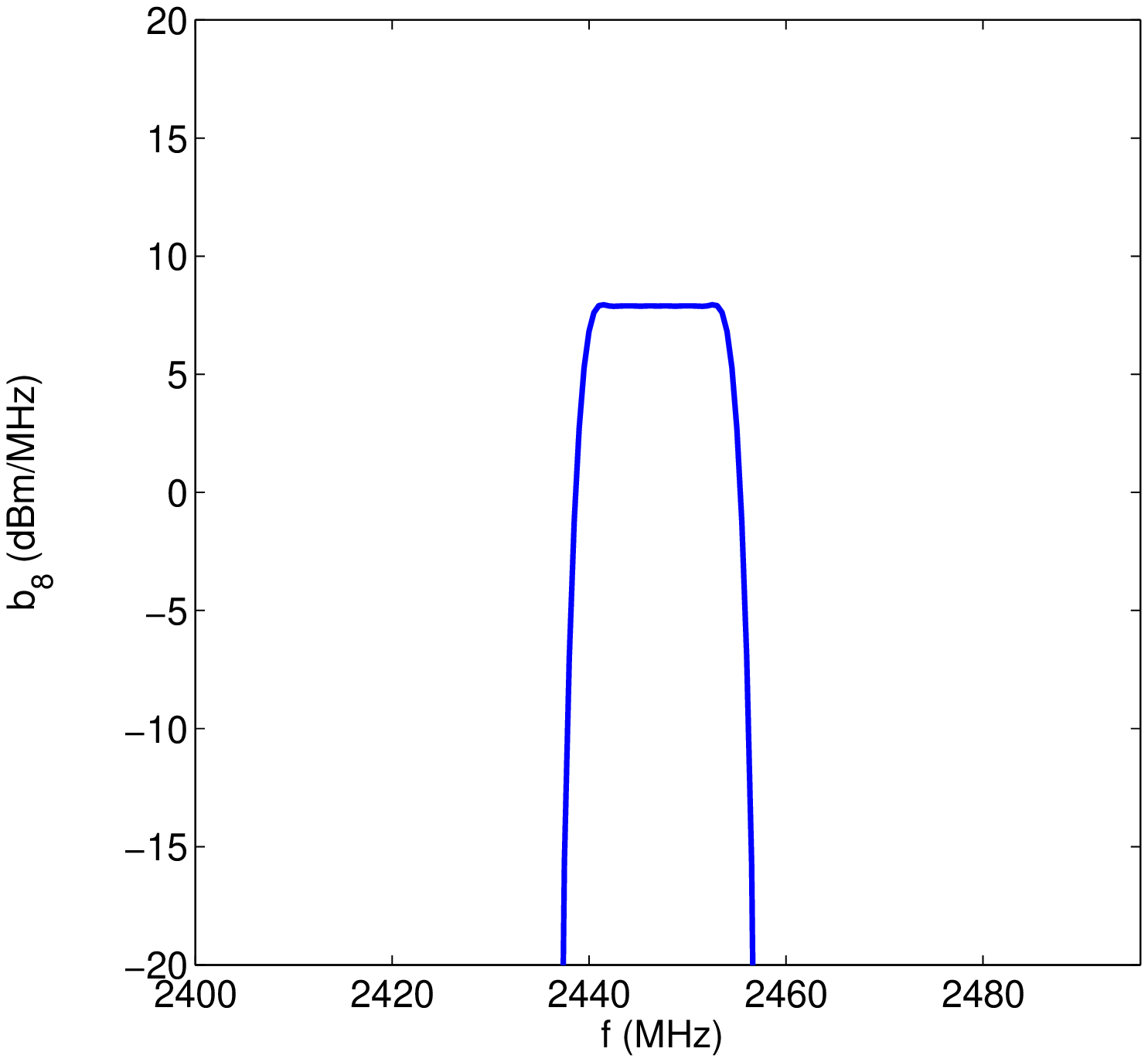}\\
\includegraphics[width=2.9cm]{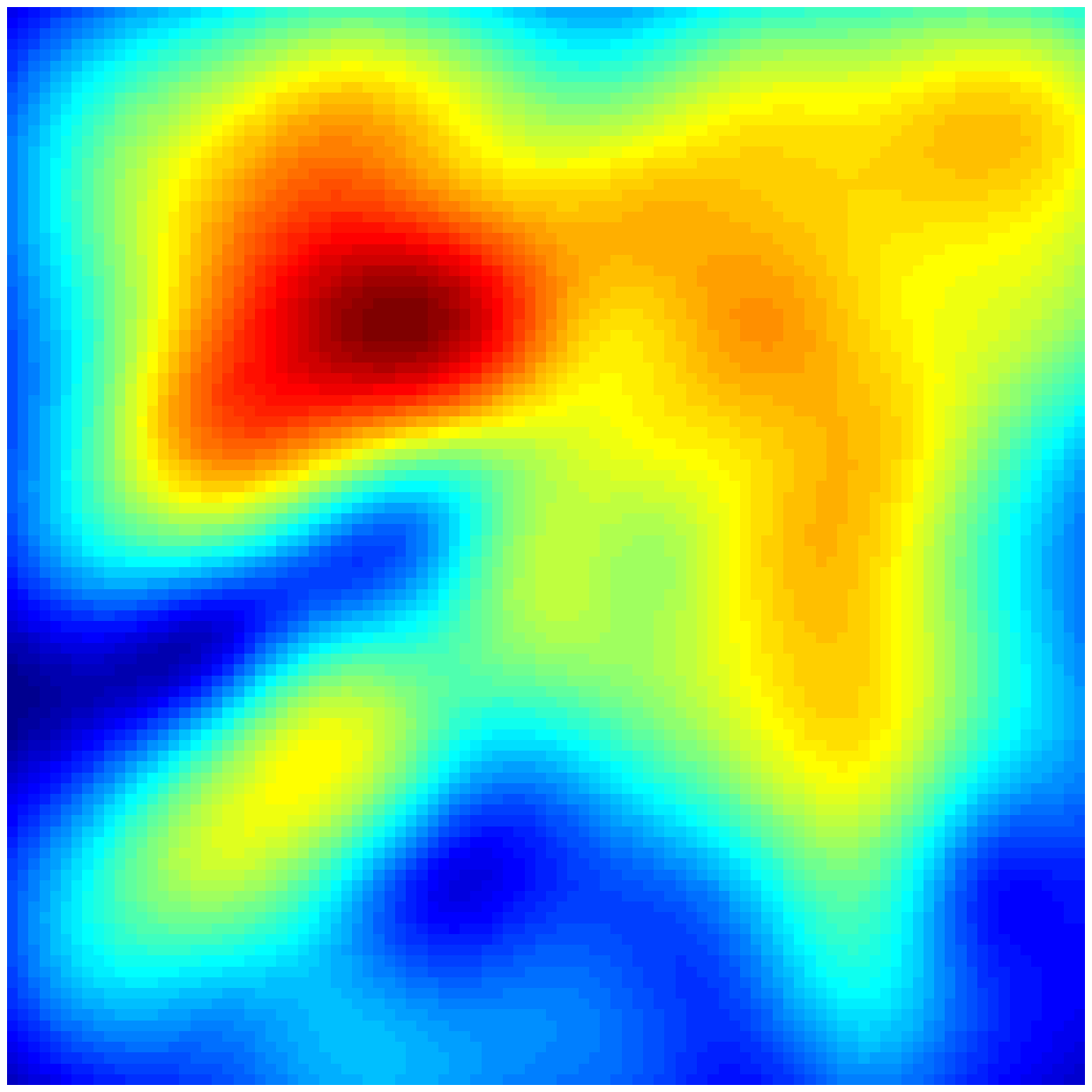}
\includegraphics[width=2.9cm]{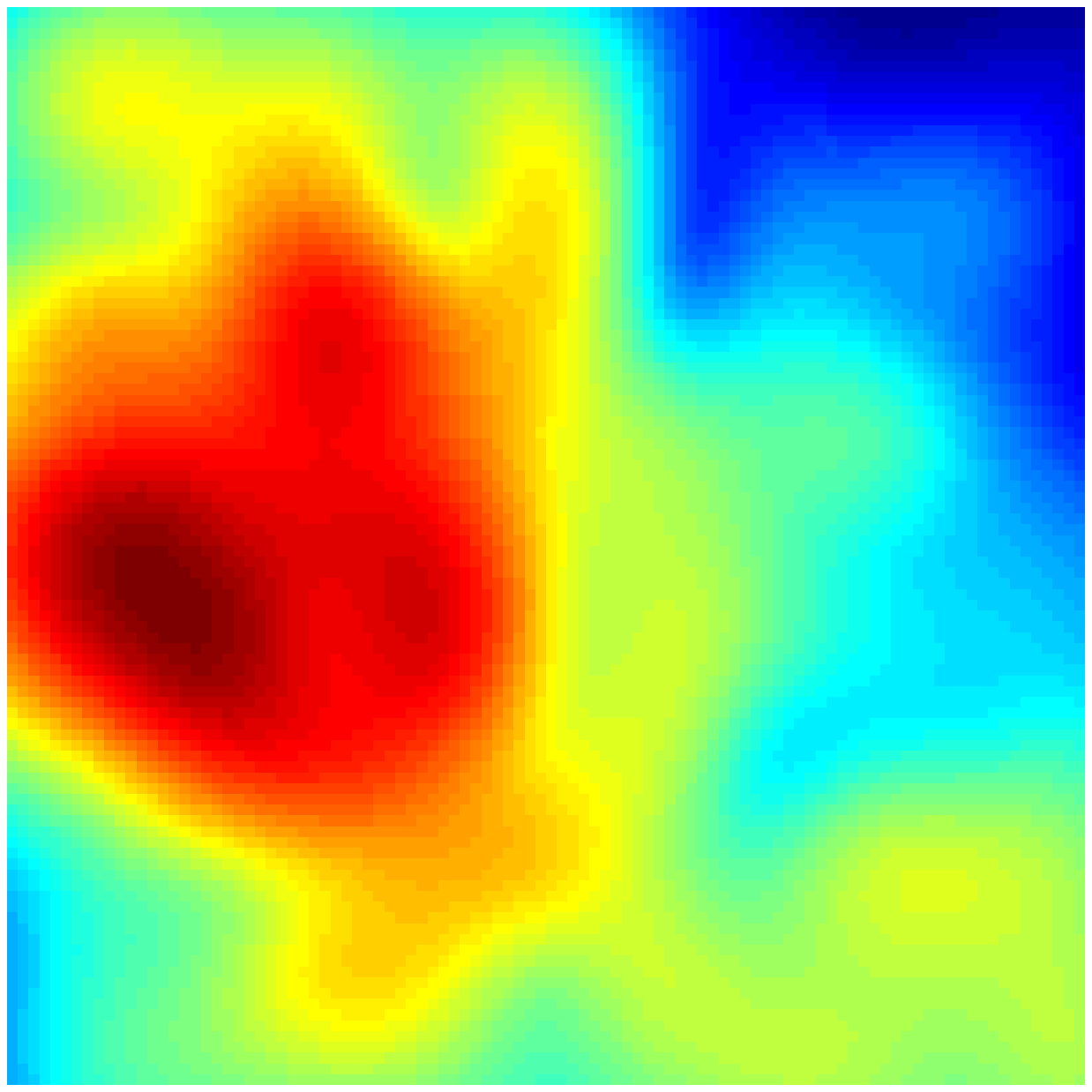}\\
\includegraphics[width=2.9cm]{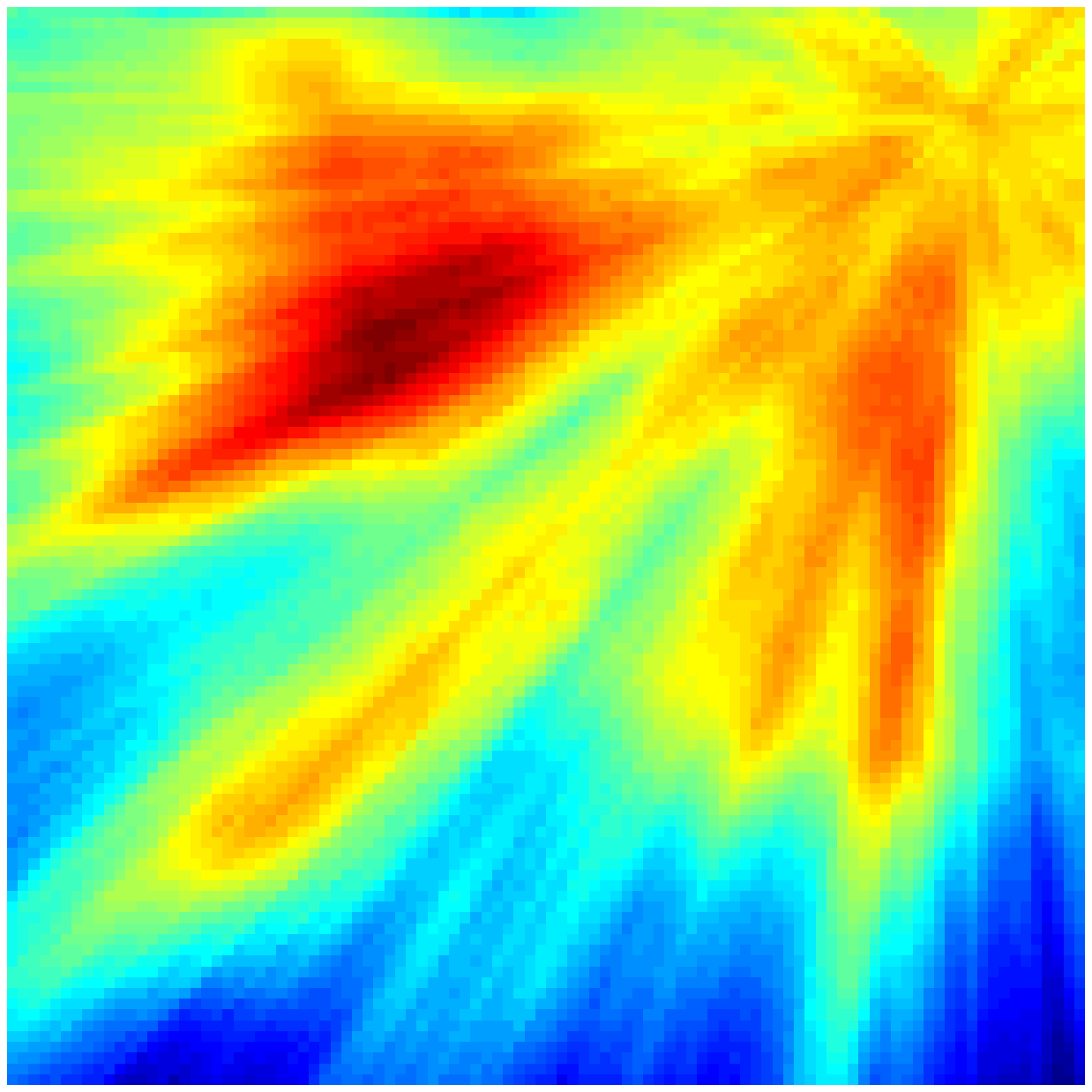}
\includegraphics[width=2.9cm]{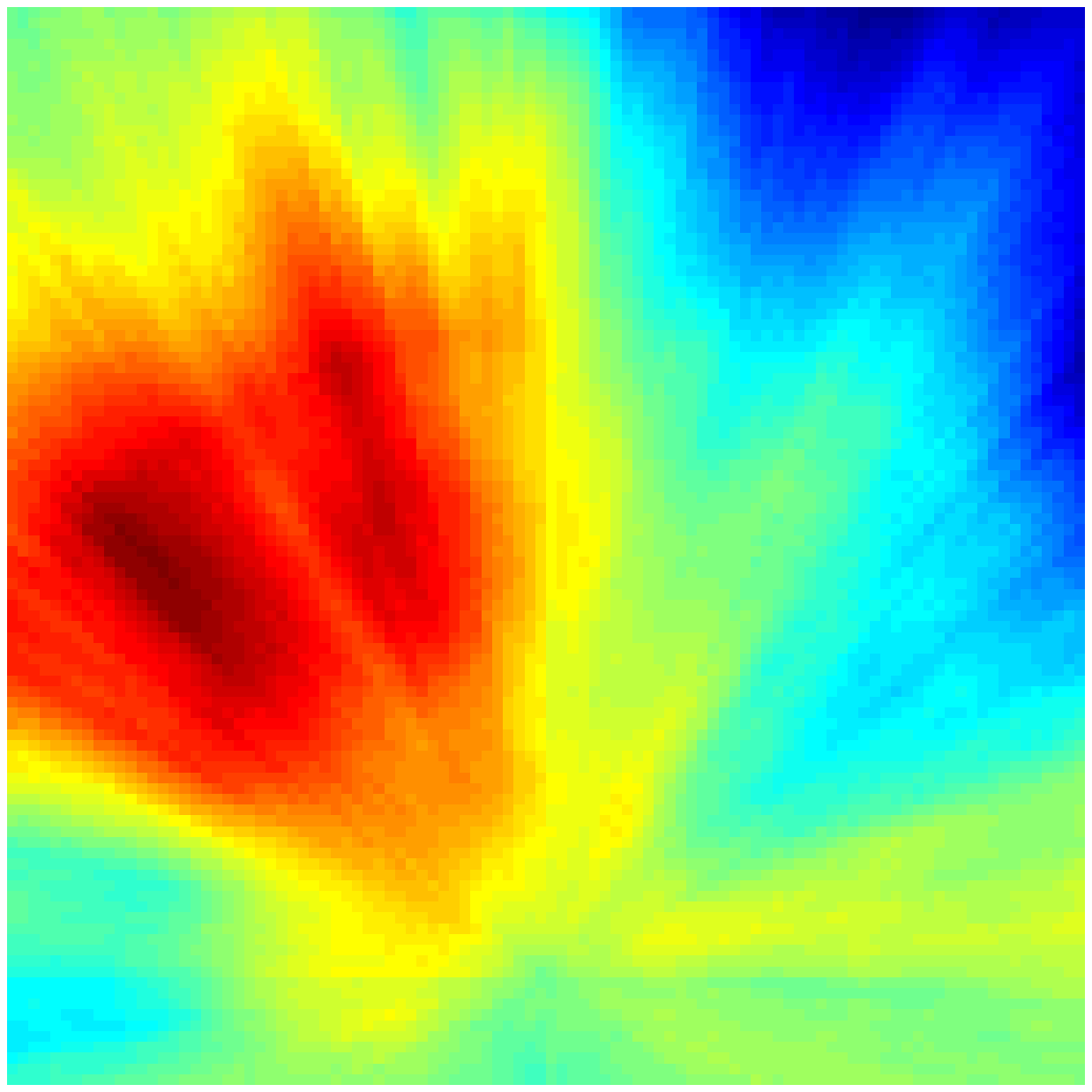}\\
\includegraphics[width=2.9cm]{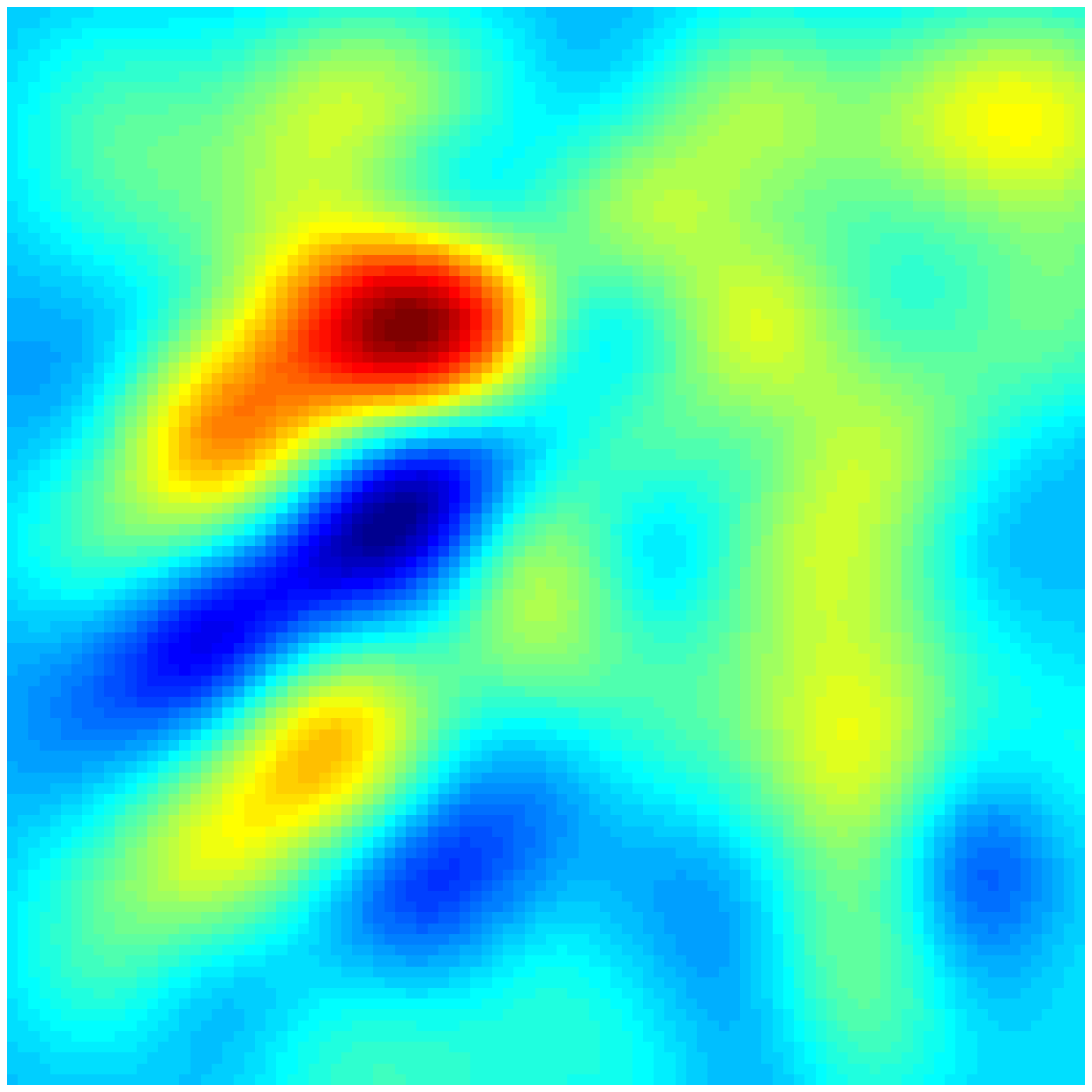}
\includegraphics[width=2.9cm]{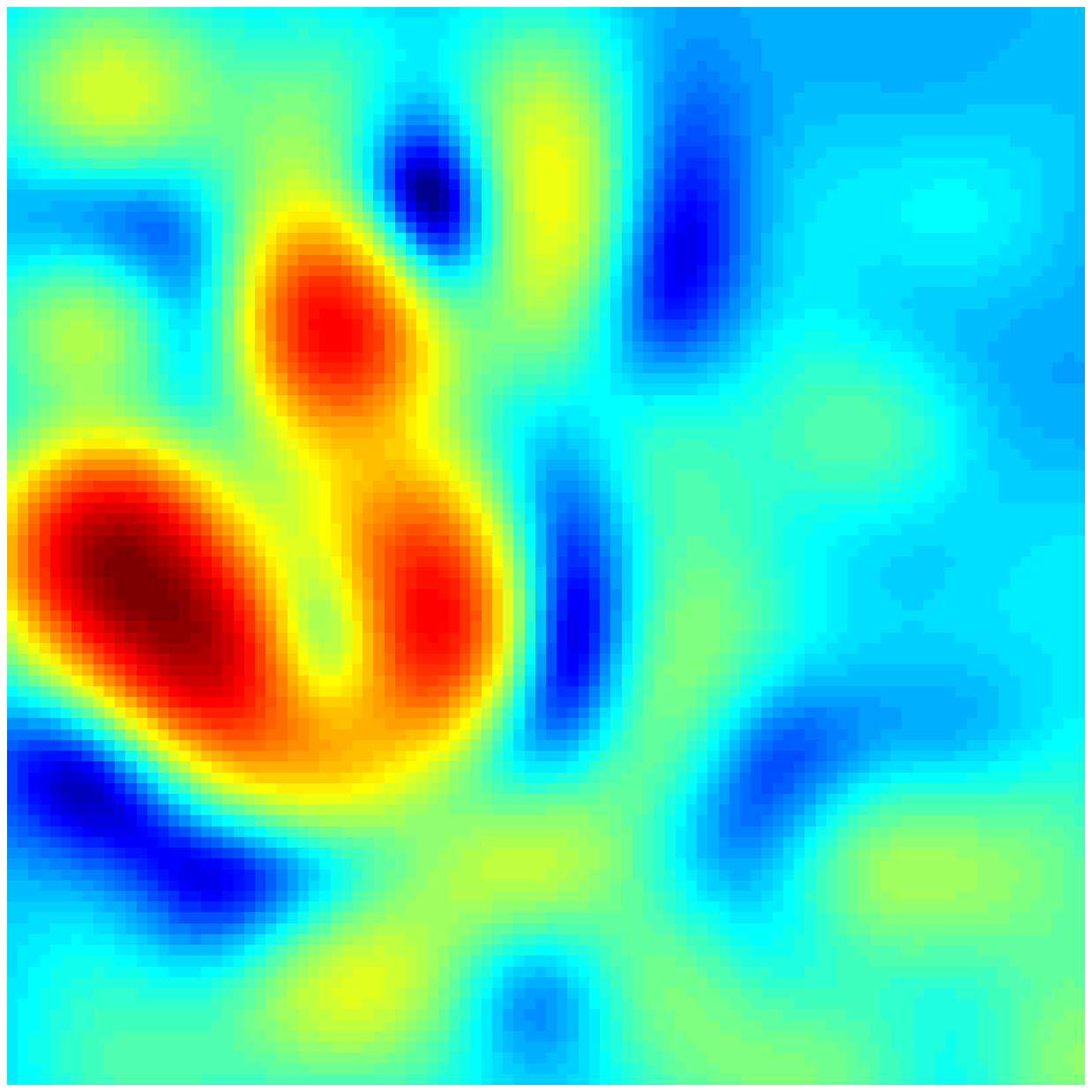}\\
\includegraphics[width=2.9cm]{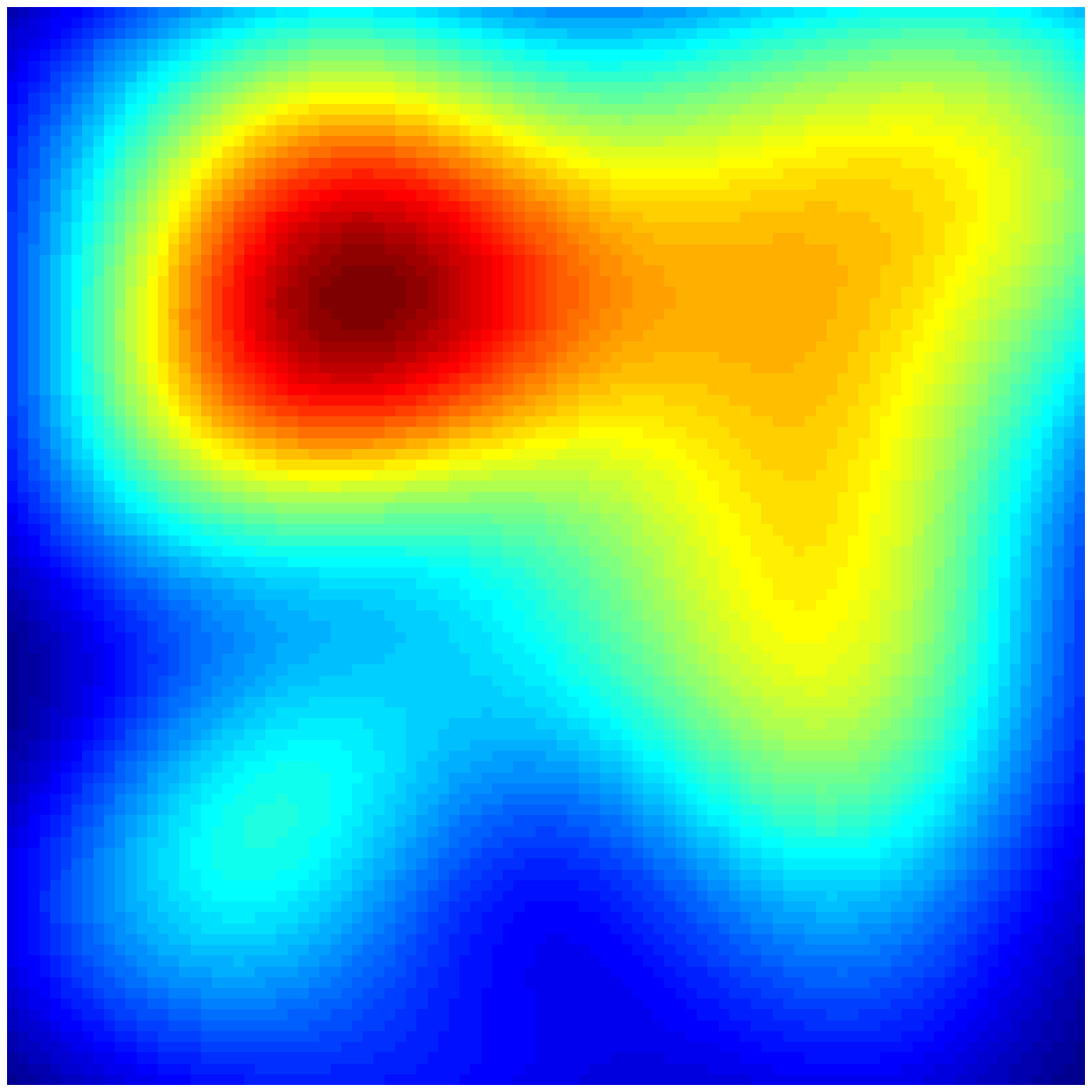}
\includegraphics[width=2.9cm]{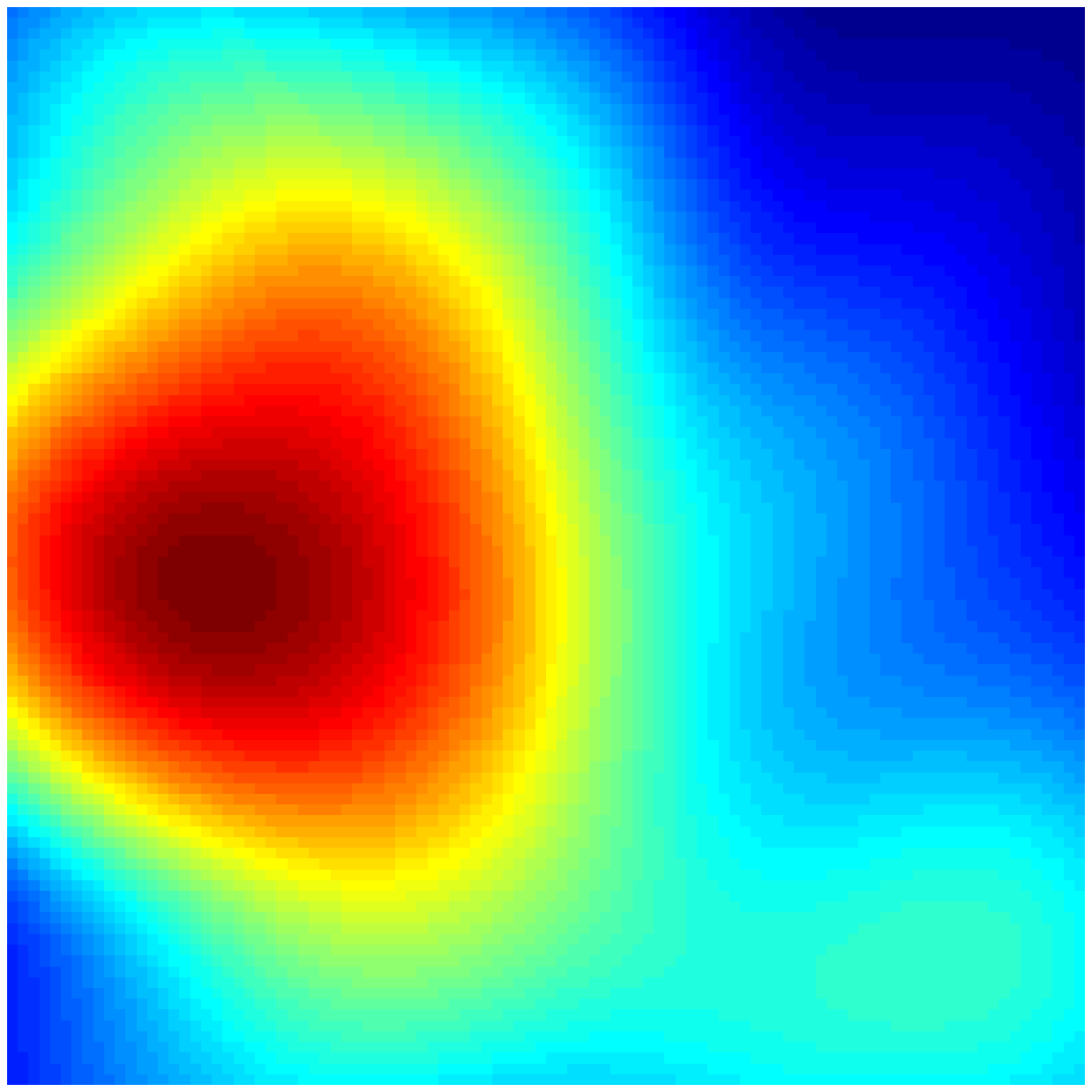}
\caption{{NBP for spectrum cartography using MKL.}}
\label{fig:nbp}
\end{figure}

Solving \eqref{nbp}  generates the PSD maps of Fig. \ref{fig:nbp}. Only $\bm\gamma_5$ and $\bm\gamma_8$  in the solution to \eqref{group_lasso_nbp} take nonzero values (more precisely $\bm\gamma_{5r}$ and $\bm\gamma_{8r},\ r=1,2$ in the MKL adaptation of \eqref{group_lasso_nbp}), which correctly reveals which frequency bands are occupied as shown in Fig. \ref{fig:nbp} (first row). The estimated PSD across space  is depicted in Fig.  \ref{fig:nbp} (second row) for each band respectively, and compared to the ground truth depicted in Fig. \ref{fig:nbp} (third row). The multi-resolution components  $c_{5r}(x)$  and $c_{8r}(x)$ are depicted in Fig. \ref{fig:nbp} (last two rows), demonstrating how kernel $k_1$ captures the coarse  pathloss distribution, while $k_2$ refines the map by revealing locations affected by shadowing.
%
%

These results demonstrate the usefulness of model $\eqref{bilinear}$  for collaborative spectrum sensing, with bases abiding to \cite{IEEE802} and multi-resolution kernels. The sparse nonparametric estimator \eqref{nbp} serves the purpose of revealing the occupied frequency bands,  and capturing the PSD map across space  per source. Compared to the spline-based approach in \cite{BMG11}, the MKL adaptation of \eqref{nbp} here provides the appropriate multi-resolution capability to capture pathloss and shadowing effects when interpolating the data across space.

\subsection{Completion of Gene Expression Data via Blind NBP}
\begin{figure}[ht]
\centering
\includegraphics[width=1.6cm]{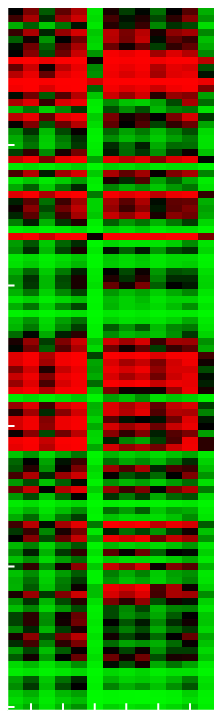}
\includegraphics[width=1.6cm]{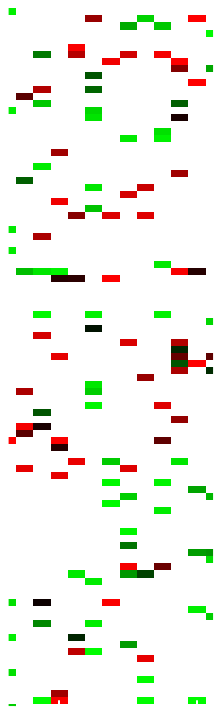}
\includegraphics[width=1.6cm]{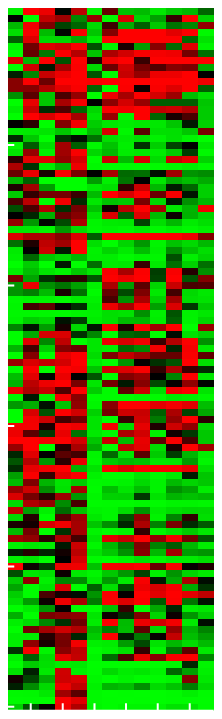}
\includegraphics[width=1.6cm]{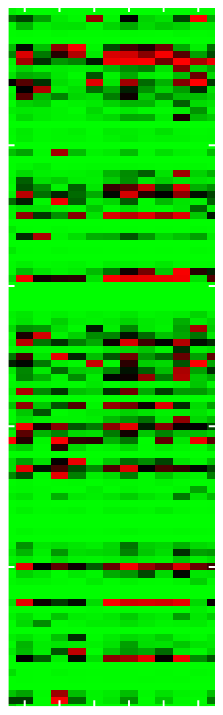}
\caption{{Microarray data completion; from left to right: original sample; $10\%$ available data; recovery via NBP; and recovery via nuclear-norm regularized LS.}}
\label{fig:microarrays}
\end{figure}

The imputation method \eqref{nuclear_norm_delta} is tested here on microarray data described in \cite{SSB04}. Expression levels of yeast across $N_g=4,772$ genes sampled at $N=13$ time points during the cell cycle  are considered. A subset of $M=100$ genes is extracted and their expression levels are organized in the matrix $\mathbf Z\in\mathbb R^{M\times N}$ depicted in Fig. \ref{fig:microarrays} (left). Severe data losses  are simulated by discarding $90\%$ of the entries of $\bZ$, including the nearly $5\%$ actually  missing data.

According to the Bayesian model \eqref{prior}, it follows that
\begin{align}
E[\bZ\bZ^T]&=\theta\mathbf R_C+\sigma^2_e \mathbf I,\ ~~
 E[\bZ^T\bZ]=\theta\mathbf R_B+\sigma^2_e \mathbf I~.\label{RZZ}\end{align}
To study  the effect of hydrogen peroxide on the cell cycle arrest,  two extra microarray datasets  $\bZ^{(1)},\ \bZ^{(2)}\in\mathbb R^{M\times N}$, synchronized with $\bZ$, are  collected in \cite{SSB04}. These two matrices are employed  to form an estimate   of $ E[\bZ\bZ^T]$, which is used instead of   $\mathbf R_C$ in \eqref{bayesian_nmc} after neglecting the noise term in \eqref{RZZ}. Since the presence of hydrogen peroxide in samples $\bZ^{(1)}$ and $\bZ^{(2)}$ induces cell cycle arrest,  the correlation between samples across time  in  $\bZ^{(1)}$ and $\bZ^{(2)}$ is altered, and thus these samples are not appropriate for estimating $E[\bZ^T\bZ]$. Alternatively, the sample estimate of $E[\bZ^T\bZ]$ is formed with the microarray data of the $(N_g-M)\times N$ genes set aside, and then used in place of $\mathbf R_{B}$ in \eqref{bayesian_nmc}.

\begin{figure}[ht]
\hspace{-0.6cm}\includegraphics[width=1\linewidth,height=0.6\linewidth]{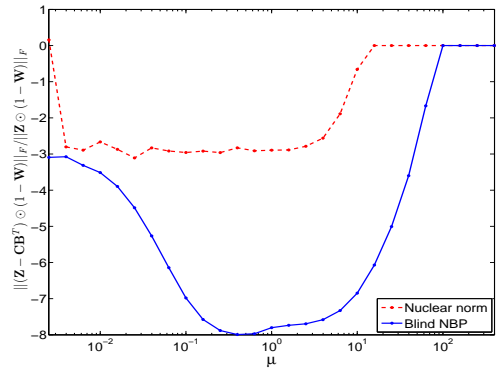}
\caption{{Relative recovery error in dB with $90\%$ missing data; comparison between blind NBP (KMC) and nuclear norm regularization. }}
\label{fig:microarray_curves}
\end{figure}

Solving \eqref{bayesian_nmc} with the available data ($10\%$ of the total) as shown in Fig. \ref{fig:microarrays} (second left) results in the matrix $\mathbf{\hat  Z}$ depicted in Fig. \ref{fig:microarrays} (second right), where the  imputed missing data introduce an average recovery error of $-8$dB [cf. Fig. \ref{fig:microarray_curves}].  In producing $\mathbf{\hat Z}$, the smoothing capability of \eqref{nuclear_norm_delta} to recover completely missing rows of  $\bZ$ (amounting to 25 in this example) is corroborated. Missing rows  cannot be recovered by nuclear norm regularization alone [cf.  \eqref{nuclear norm}], even if  $\bZ$ is padded with expression levels of the discarded $N_g-M$ genes. Fig. \ref{fig:microarrays} (right) presents this case confirming that its performance dagrades w.r.t. NBP; while  Fig. \ref{fig:microarray_curves} illustrates the sensitivity of the estimation error  to the cross-validated regularization parameter $\mu$ for both estimators.
Similar degraded results are observed when imputing missing entries of $\bZ$ using the impute.knn() and svdImpute() methods, as implemented in the R packages pcaMethods and BioConductor-impute. These two methods were applied to the padded $\bZ$, after the requisite discarding of the 25 missing rows,  resulting in recovery  errors on the remaining missing entries at $-3.84$dB and  $-0.12$dB (with parameter nPcs$=12$), respectively.

\subsection{Network Flow Prediction via Blind NBP}
\begin{figure}[ht]
\centering
\includegraphics[width=0.6\linewidth]{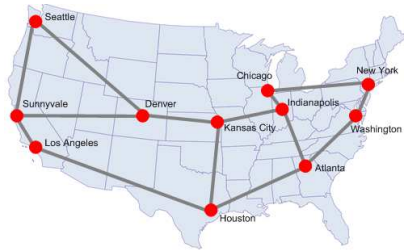}
\caption{Internet 2 network topology graph \cite{Abilene}.}
\label{fig:abilene_network}
\end{figure}

The Abilene network in Fig. \ref{fig:abilene_network}, a.k.a. Internet 2, comprising $11$ nodes and $M=30$ links \cite{Abilene}, is utilized as a testbed for traffic load prediction.
Aggregate link loads $z_{mn}$ are recorded every $5$ minute intervals in the morning of December 22, 2008, between  12:00am and 11:55pm, and are collected in the first $N/2=144$ columns of matrix $\bZ\in \mathbb R^{M\times N}.$
These samples are then used to predict link loads hours ahead, by capitalizing on their mutual cross-correlation, the periodic correlation  across days, and their interdependence across links as dictated by the network topology.

\begin{figure}
\centering
\includegraphics[width=0.6\linewidth]{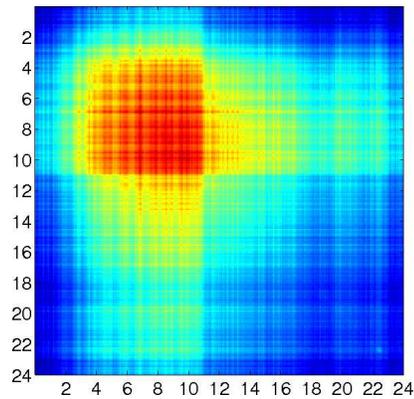}
\caption{ Sample estimates of  $E(\bZ\bZ^T)$ for link loads across time, are used to replace   $\mathbf R_C$ and   $\Ky$. }
\label{fig:correlation_abilene}
\end{figure}

\begin{figure}
\centering
\includegraphics[width=0.9\linewidth]{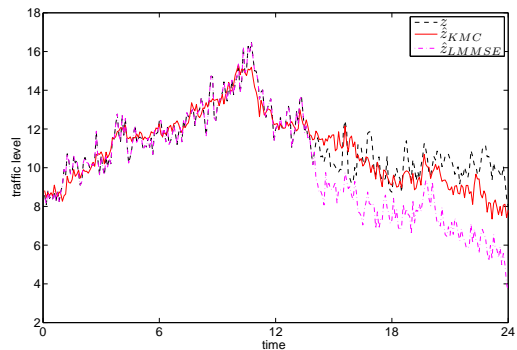}
\caption{{ Network prediction via KMC (blind NBP). Measured and predicted traffic on link $m=21$.}}
\label{fig:network_prediction}
\end{figure}

The correlation matrix $E(\bZ\bZ^T)$ represented in Fig. \ref{fig:correlation_abilene}  is estimated with training samples collected during the two previous weeks, from December 8 to December 21, 2008, and substituted for $\mathbf R_C$ in \eqref{bayesian_nmc} according to \eqref{RZZ}. A singular point at 11:00am in the traffic curve, as depicted in black in Fig. \ref{fig:network_prediction}, is reflected in the sharp transition noticed in  Fig. \ref{fig:correlation_abilene}. On the other hand, $\mathbf R_B$ is not estimated but derived from the network structure. Supposing  i.i.d. flows across the network, it holds that $E(\bZ^T\bZ)=\sigma_f^2\mathbf R^T\mathbf R$, where $\mathbf R$ represents the network routing matrix and $\sigma_f^2$ the flow variance. Thus, $\sigma_f^2\mathbf R^T\mathbf R$, was used instead of  $\mathbf R_B$ in \eqref{bayesian_nmc}, with $\sigma^2_f$ adjusted to satisfy $\textrm{tr} (E[\bZ^T\bZ])=\textrm{tr}(  E[\bZ\bZ^T])$.

Fig. \ref{fig:network_prediction} shows link loads predicted by \eqref{bayesian_nmc} on December 22, 2008, for a representative link, along with the actually recorded samples for that day.
Prediction accuracy is compared in Fig. \ref{fig:network_prediction} to a base strategy comprising independent LMMSE estimators per link, which yield a relative prediction error $e_p=0.22$ aggregated across links,  against $e_p=0.15$ that results from \eqref{bayesian_nmc}. Strong correlation among samples from 12:00am to 2:00pm [cf. Fig. \ref{fig:correlation_abilene}] renders  LMMSE prediction accurate in this interval,  relying on single-link data only.  The benefit of considering the links jointly is appreciated in the subsequent interval from 2:00pm to 11:55pm, where the traffic correlation with morning samples fades away and the network structure  comes to add valuable information, in the form of $\mathbf R_B$,  to stabilize prediction.

%
%
%
%
%
%
\section{Summary}
A new methodology was outlined in this paper by cross fertilizing sparsity-aware signal processing tools with
kernel-based learning. It goes well beyond translating sparse vector
regression techniques into their nonparametric counterparts, to
generate a series of unique possibilities such as kernel selection
or kernel-based matrix  completion. The
present article  contributes to these efforts by advancing NBP as the
cornerstone of sparse KBL, including blind versions that emerge
as nonparametric  nuclear norm regularization and
dictionary learning.

KBL was connected with GP analysis, promoting a Bayesian viewpoint where kernels convey prior information. Alternatively, KBL can be regarded as an interpolation toolset though its connection with the NST,  suggesting that the impact of the prior model choice is attenuated when the size of the dataset is large, especially when kernel selection is also incorporated.

All in all, sparse KBL was envisioned as a fruitful
research direction. Its impact on signal processing practice
was illustrated through a diverse set of application paradigms.

\section*{Appendix}
\subsection*{Proofs of Properties P1-P3}
 \begin{proof} 1) If white noise $n(x): x\in \mathbb R$ is fed to an ideal low-pass filter with cutoff frequency $\omega_{\max}=\pi$, then $r(\xi):=E(z(x)z(x+\xi))=\sinc(\xi)$ is the autocorrelation of the output $z(x)$. Hence, $\mathbf K$ equals the covariance matrix of $\mathbf z^T:=[z(x_1),\ldots,z(x_N)]$, and as such $\mathbf K\succeq\mathbf 0$.
 \end{proof}

\begin{proof}
 2) Rewrite the kernel $f_{x'}(x):=\sinc(x-x')$ as a function parameterized by $x'$. Then, the NST applied to the bandlimited $f_{x'}(x)$ yields $f_{x'}(x)=\sumnz f_{x'}(n)\sinc(x-n)=\sumnz\phi_n(x')\phi_n(x)$.
 \end{proof}

\begin{proof}
3) Upon defining  $\alpha_n:=f(x_n)$, the reconstruction formula $f(x):=\sumnz f(n)\sinc(x-n)$ gives the kernel expansion of  $f\in\Bpi$. Hence, by definition of the RKHS norm  $\|f\|^2_{\calHx}=\sumnz \sum_{n'\in\mathbb Z} f(n)\sinc(n-n')f(n')$. Substituting the reconstructed $f(n)=\sum_{n'\in\mathbb Z}\sinc(n-n')f(n')$ into the last equation yields $\|f\|_{\calHx}^2=\sumnz f^2(n)$.
 \end{proof}

\subsection*{Design of Algorithm 1}
In order to rewrite  the cost $\frac{1}{2}\|(\bZ-\bC\bB^T)\odot\bW\|_F^2+\frac{\mu}{2} \left[\textrm{Tr}(\bC^T\Kx^{-1}\bC)+\textrm{Tr}(\bB^T\Ky^{-1}\bB)\right]$ in terms  $\bci=\bC\bei$ and $\bbi=\bB\bei$, representing  the $i$-th columns of matrix $\bB$ and $\bC$, respectively,  define $\bbCi=\bC-\bci\bei^T$ and decompose
$\bC\bB^T=\bbCi\bB^T+\bci\bbi^T$. Then rewrite the cost as
\begin{align}
&\frac{1}{2}\|(\bZ_i-\bci\bbi^T)\odot\bW\|_F^2+\frac{\mu}{2}\bci^T\Kx^{-1}\bci\label{app:cost_ci}
\end{align}
after defining $\bZ_i:=\bZ-\bbCi\bB^T$ and discarding regularization terms not depending on $\bci$.

Let $\vv(\bW)$ denote the vector operator that concatenates columns of $\bW$,  and $\mathbf D:=\dd[\mathbf x]$ the diagonal matrix operator such that $d_{ii}=x_i$.  The Hadamard product can be bypassed by defining $\Dw:=\dd[\vv(\bW)]$, substituting $\|\mathbf X\|_F=\|\vv(\mathbf X)\|_2$, and using the following identities
 \begin{align}
 \vv(\bW \odot \mathbf X)&=\Dw\vv(\mathbf X),\nonumber\\
    \vv(\mathbf X_i\bbi^T)&=(\bbi\otimes \mathbf I_M)\vv(\mathbf X_i)\label{app:vec_otimes}
  \end{align}
     with $\otimes$ representing the Kroneker product. Applying \eqref{app:vec_otimes} to  \eqref{app:cost_ci} yields
\begin{align}
\frac{1}{2}\|\Dw\vv(\bZ_i)-\Dw(\bbi\otimes \mathbf I_M)\bci\|_2^2+\frac{\mu}{2} \bci^T\Kx^{-1}\bci\label{app:cost_cidiag}
\end{align}

Equating the gradient of \eqref{app:cost_cidiag} w.r.t. $\bci$ to zero, and solving for $\bci$ it results
\begin{align}
&\bci=\mathbf H_i^{-1}   (\bbi^T\otimes \mathbf I_M)\Dw\vv(\bZ_i)\nonumber\\&\mathbf H_i:=   \bbi^T\otimes \mathbf I_M)\Dw\Dw (\bbi^T\otimes \mathbf I_M)   +\mu \Kx^{-1}  \label{app:solution_ci_cmplx}
\end{align}

It follows from \eqref{app:vec_otimes} that
$(\bbi^T\otimes \mathbf I_M)\Dw\vv(\bZ_i)=(\bW\odot \bZ_i)$, and it can be established  by inspection that
$(\bbi^T\otimes \mathbf I_M)\Dw\Dw (\bbi^T\otimes \mathbf I_M)=\sum_{n=1}^N b_{in}^2 \dd[\mathbf w_n]=\dd\left[ \bW (\bbi\odot\bbi)\right]$, so that \eqref{app:solution_ci_cmplx} reduces to $\bci= \left( \dd\left[ \bW (\bbi\odot\bbi)\right]+\mu\Kx^{-1}\right)^{-1}(\bW\odot \bZ_i)\bbi$,  coinciding with  the update for $\bci$ in Algorithm 1. The corresponding update for $\bbi$ follows from parallel derivations.

\end{document}